\documentclass[man]{apa6}

\usepackage{bm}
\usepackage{amsmath}
\usepackage{verbatim}
\usepackage{algorithm}
\usepackage{algpseudocode}
\usepackage{booktabs}
\usepackage{graphicx}
\usepackage{setspace}
\usepackage{multirow}
\usepackage{listings}
\usepackage[natbibapa]{apacite}

\usepackage{natbib}

\usepackage[usenames,dvipsnames]{color}    
\DeclareCaptionLabelSeparator*{spaced}{\\[2ex]}
\AtBeginEnvironment{tabular}{\singlespacing}

\title{metboost: Exploratory regression analysis with hierarchically clustered data}
\shorttitle{metboost}
\author{Patrick J. Miller$^1$, Daniel B. McArtor$^1$, Gitta H. Lubke$^{1, 2}$}
\affiliation{$^1$ University of Notre Dame, $^2$ VU University Amsterdam}
\authornote{ Patrick J. Miller expresses his appreciation to Daniel B. McArtor for his critical insight that motivated the method.\\
Gitta H. Lubke is supported by NIDA R37 DA-018673. The computational work was done on clusters acquired through NSF MRI BCS-1229450. \\
Correspondence concerning this abstract should be addressed to Patrick J. Miller, 110 Haggar Hall, University of Notre Dame, Notre Dame, IN 46656. Email: pmille13@nd.edu.}

\begin{document}

\abstract{As data collections become larger, exploratory regression analysis becomes more important but more challenging. When observations are hierarchically clustered the problem is even more challenging because model selection with mixed effect models can produce misleading results when nonlinear effects are not included into the model (Bauer and Cai, 2009). A machine learning method called boosted decision trees (Friedman, 2001) is a good approach for exploratory regression analysis in real data sets because it can detect predictors with nonlinear and interaction effects while also accounting for missing data. We propose an extension to boosted decision decision trees called {\em metboost} for hierarchically clustered data. It works by constraining the structure of each tree to be the same across groups, but allowing the terminal node means to differ. This allows predictors and split points to lead to different predictions within each group, and approximates nonlinear group specific effects. Importantly, metboost remains computationally feasible for thousands of observations and hundreds of predictors that may contain missing values. We apply the method to predict math performance for 15,240 students from 751 schools in data collected in the Educational Longitudinal Study 2002 (Ingels et al., 2007), allowing 76 predictors to have unique effects for each school. When comparing results to boosted decision trees, metboost has 15\% improved prediction performance. Results of a large simulation study show that metboost has up to 70\% improved variable selection performance and up to 30\% improved prediction performance compared to boosted decision trees when group sizes are small.}


\maketitle

It can be challenging to make predictions and find structure in hierarchically clustered data with many predictors. Hierarchical clustering is frequently observed in psychology. For example, individuals can be nested within schools, organizations, regions, or families. Many studies initially consider a large number of variables. This could range from tens of thousands to simply a large number that is inconvenient to include in a parametric model simultaneously (e.g. 50 or 100). For instance, data might be collected using sensors or computer logs, collected from virtual environments like websites or online textbooks, or compiled from large publicly funded surveys. In these studies, research questions are often broad, specifying only which variable is the outcome and which variables are plausible predictors. Initially the goal is the data exploration: understanding how the predictors relate to the outcome while making as few assumptions as possible \citep{tukey_analyzing_1969}. Using a part of the data for exploration can suggest further research questions, testable hypotheses, and appropriate models to fit to the data. It can also enhance our ability to make predictions for student learning, depression risk, or employee well-being. 

While the benefits of exploratory regression analysis are well known, many intuitive approaches require strong assumptions or do not scale to a large number of variables. Common methods are plots (scatter plots, regression diagnostics), estimating pearson correlations between the outcome and all predictors, and stepwise or best subsets regression. While data and model visualization can highlight nonlinear effects, these approaches do not work well with a large number of predictors. Linear models (pearson correlation, stepwise regression) can work with a large number of predictors but require assuming that the relationship between the outcome and predictors is linear. Thus linear models can fail to detect predictors with nonlinear and interaction effects unless these terms are included a priori. Furthermore, stepwise and best subsets regression are also not guaranteed to find the best possible model, even if all effects are linear \citep{thompson_stepwise_1995}. Finally, none of these approaches explicitly account for hierarchical clustering.

Decision trees are a compelling alternative for exploratory regression analysis because they allow an arbitrary large number of predictors, can approximate nonlinear effects without specification, and handle missing data. A decision tree is a set of binary decision rules built in a data driven way. For example, a decision tree predicting math score on a standardized test might first split on the variable GPA $< 3.2$. Within each of the two resulting regions, (GPA $< 3.2$, GPA $\geq 3.2$), the search through all predictors is carried out again. Within one region, a split might be found on homework completion rate $ < 80\%$. The predictions of the tree are the mean math performance in each region. In this region, for example, the prediction would be the math performance of students with GPA $< 3.2$ and homework completion rate $> 80\%$. The problem with a single decision tree is that the structure of the trees can vary considerably from sample to sample (i.e. it has high variance). So while interpretation of the tree as a set of decision rules is straightforward, the high sample to sample variability of any individual tree does not instill confidence in any given interpretation. For example, fitting a tree in a different sample could result in homework completion being the first splitting variable, which potentially leads to different subsequent splits and ultimately a completely different conceptual interpretation of the tree.
 
{\em Boosted decision trees}, or boosting, is a method that reduces the variance of individual decision trees by combining the predictions of many trees in an ensemble. In boosting, each tree is fit by increasing the weight of observations that are most poorly predicted by the previous trees in the model. In our example, the first tree with splits on GPA and homework completion rate would not explain variability in math performance due to additional factors like teacher expectations or math self efficacy. Using boosting, these predictors could be included in subsequent trees that are fitted to the residuals of the previous trees. The resulting ensemble has higher prediction accuracy and less variability than individual trees, but is no longer easy to interpret as a set of decision rules. Instead, boosting models are typically interpreted by ranking how often each predictor is used for splitting and how effective the corresponding splits were. In our example, GPA might rank highest among all the variables considered. Nonlinear effects can be detected by examining plots of the effects of individual predictor effects implied by the model. 

Generally, a boosted decision tree ensemble can be viewed as an approximation of a highly complex or nonlinear function relating many predictors to an outcome. This property is ideal for exploratory regression analysis because it allows exploration of the regression relationship between an outcome and many predictors while making few assumptions about the structure of this relationship. However, it is not clear how best to incorporate hierarchically clustered data in a boosted decision tree ensemble.

In psychology, observations are often hierarchically clustered into groups. We refer to the variables that induce clustering within observations simply as {\em grouping variables}: school, family, organization, team, or therapist are some common examples. Throughout this paper, we will use a data set in education as a concrete example of these hierarchical structures, with students clustered within schools. When students are clustered within schools, the effect of school is commonly modeled as a random effect in a mixed effect model \citep{harville_maximum_1977, henderson_best_1975}. A mixed effects model defines a set of fixed effects which are the same across all groups, and a set of random effects that are unique to each group \citep{laird_random-effects_1982}. The regression intercept and slope for certain predictors are often allowed to vary by group, which allows each group to have unique predictor effects. The levels of a grouping variable are assumed to be sampled randomly from a population of all possible levels. For example, the observed schools in the sample are a random sample of all schools in a state. Accounting for the random effect of the grouping variable corrects the standard errors of parameter estimates used in hypothesis tests, which are overly liberal if the clustering is ignored \citep{henderson_best_1975}. 

For prediction, the advantage of a mixed-effect models is that they provide shrunken (or empirical bayes) estimates of the random effects \citep{skrondal_prediction_2009}. For a random intercept model, these estimates are closer to the within-group mean or grand mean depending on the within-group variance and the sample size within each group \citep{gelman_data_2006,afshartous_prediction_2005}. In finite samples, shrunken or empirical bayes estimates of the random effects are more efficient than simply using the group means alone \citep{afshartous_prediction_2005}.

A natural approach for exploratory regression analysis with large mixed effect models is to fit a set of alternative models followed by model selection. However, model selection is problematic because it is difficult to both detect and disambiguate nonlinear predictor effects from random predictor effects. Others have recommended allowing all predictors to have group specific effects \citep{barr_random_2013}. However, such a strategy is statistically and computationally problematic with finite samples. Critically, it has also been shown that nonlinear effects not included in the model can be incorrectly approximated by a random slope \citep{bauer_consequences_2009}. Thus, exploratory regression analysis with mixed effect models by model selection can result in an incorrect final model with a misleading interpretation. 

As an alternative to model selection, we propose a procedure specifically designed for exploratory regression analysis with mixed effect models. This is done by extending boosted decision trees to take hierarchical clustering into account. We call the procedure \textit{metboost} (mixed effects tree boosting) because it combines boosted decision trees with mixed effects models.  Our approach is implemented in the R package {\em mvtboost} \citep{miller_finding_2016} and is freely available on CRAN. Building on boosted decision trees capitalizes on their flexibility for exploratory data analysis and their ability to detect nonlinear effects without specification, while simultaneously handling missing data in the predictors. The resulting model can still be interpreted by ranking predictors by their contribution to the model and by plotting their model-implied effects. Combining boosted decision trees with mixed effects models improves the efficiency of predictions because the predictions are shrunken according to group size. For example, the predicted values for a school with a small number of students are shrunken toward the average across all schools. Furthermore, the interpretation of metboost is improved because the levels of grouping variable are treated as a sample of all possible levels (e.g. a sample of all schools in a state).

The essence of our approach is to constrain the split points of each tree to be the same across groups, but to allow the terminal node means of the tree (the predictions) to vary by group. In the education example, the tree split point GPA $< 3.2$ is the same across schools, but the math performance in each region is specific to each school and depends on the distribution of GPA within the school. This approach is implemented in metboost by representing each tree in the ensemble as a fully observed design matrix of indicator variables denoting terminal node membership. Each column in the new design matrix is allowed to vary by group using the R package {\em lme4} \citep{bates_fitting_2015}. This novel representation allows metboost to efficiently explore hierarchically clustered data with many predictors while handling missing data. However, incorporating random effects with boosted decision trees does incur computational costs, with an analysis taking on the order of minutes for medium data sets (1K observations) or hours for large data sets (>10K observations). 

Many parametric, semi-parametric, and recursive partitioning approaches have been proposed in the last 10 years for exploratory regression analysis with mixed effect models. Exploratory parametric methods penalize the random and/or fixed parameters of linear mixed models so that the number of parameters can be larger than the number of samples \citep{fan_variable_2012, ni_variable_2010, schelldorfer_glmmlasso:_2014, schelldorfer_estimation_2011, rohart_selection_2014}. Similarly, regularization of linear mixed models with many parameters can also be achieved by updating coefficients of the linear mixed effect model stepwise through likelihood based boosting \citep{tutz_generalized_2010,groll_variable_2014}. The primary drawback to these approaches for exploratory regression analysis is that all effects are assumed to be linear. 

To account for nonlinear effects, several semi-parametric methods combine additive models and mixed effect models to capture nonlinear effects with splines \citep{wood_generalized_2006}. Two examples include the \texttt{gamm} function in the R package {\em mgcv}  \citep{wood_generalized_2006}, which is based on {\em nlme} \citep{pinheiro_mixed-effects_2006}, and the \texttt{gamm4} function  \citep{wood_gamm4:_2014}  based on the R package lme4. While these approaches can capture nonlinear effects,  these models can be slow to estimate with a large number of predictors and samples.

Recently, recursive partitioning approaches have been used to capture nonlinear fixed effects. These approaches are faster to estimate with a large number of predictors. Examples include fitting linear mixed models within each region of a decision tree \citep{fokkema_detecting_2015}, and modeling fixed effects with a decision tree while accounting for random effects with a linear mixed model in a separate step \citep{hajjem_mixed-effects_2014, sela_re-em_2012}. The interpretability of the resulting trees is helpful for exploratory regression analysis, but requires specification of the mixed effects model a priori. The most recent approach uses splines to allow group specific nonlinear effects for a single covariate in boosted decision trees \citep{pande_boosted_2016}. However, only a single covariate can have group specific effects. 

Compared to existing parametric, semi-parametric and recursive partitioning procedures, metboost is a more general solution for exploratory regression analysis with many predictors in the presence of hierarchical clustering. Our approach is useful because it has all of the following features:

\begin{itemize}
\item Can be used for selection of predictors with random effects without a priori model specification or model selection.
\item Allows group-specific (non)linear effects to be detected and visualized without including additional terms in the model a priori.
\item Provides a built-in imputation procedure without a separate imputation step.
\item Is computationally feasible with several hundred predictors and thousands of observations.
\end{itemize}
In some cases, all four goals may not be necessary for exploratory regression analysis (e.g. if the number of predictors is small, or the model is already known). In this case, one or more procedures described above will be more appropriate and interpretable for exploratory regression analysis than metboost. 

The only other method that has all four features are tree ensembles (such as boosted decision trees) with the grouping variable included as a candidate for splitting. This effectively treats the grouping variable as fixed rather than random. Treating the grouping variable as fixed is appropriate if all possible levels of the variable are observed in the sample, group sizes are large, and the research questions address the specific levels observed in the sample. In education, this might be true if the number of schools of interest is finite and all appear in the sample. However, treating the grouping variable as fixed is not appropriate when the research goal is to make predictions for observations at unobserved levels of the grouping variable, or to identify important predictors while generalizing to the population of possible levels of the grouping variable. In this case, the grouping variable should be treated as random, e.g. if the schools in the sample are drawn from a population of schools in a state. Importantly, treating the grouping variable as fixed does not make the most statistically efficient predictions when group sizes are small. The critical difference between metboost and boosted decision trees is that the effect of the grouping variable is incorporated in every tree in metboost with predictions shrunken proportional to group size, but chosen in a data driven way and unshrunk in boosting. 

While metboost is good for exploratory regression analysis, there are two primary limitations. The first is that while metboost is more general than many of the existing approaches, it is less interpretable than (semi)-parametric models. However, when the research questions are exploratory rather than confirmatory, this tradeoff is useful because it informs the appropriate model to fit to the data in a second step. The second limitation to metboost (and many other machine learning methods) is that using the method effectively for variable selection and prediction requires finding optimal values for meta-parameters: the number of trees, the step size, and tree-depth. Tuning all three by cross-validation is critical for achieving optimal prediction accuracy and variable selection performance, but adds additional computation time and complexity. In our implementation, we made it easy to tune these models and improved computation time using parallelization.

In the rest of the paper, we describe the metboost model and how it is estimated, tuned, and interpreted. To illustrate the advantages of treating the grouping variable as random, we compare the performance of metboost to boosted decision trees in a Monte Carlo simulation. Finally, we describe an empirical example where the math ability of 15,240 students from 751 high schools is predicted from 76 variables using data from the Educational Longitudinal Study 2002 \citep{ingels_education_2007}.  We begin with a brief overview of decision trees, boosting, and mixed effect models. 

\section{Decision Trees and Boosting}

\subsection{Decision Trees}

A general way to view a decision or classification tree is as a piecewise function approximation of an unknown function of the predictors. Notationally, they can be represented as:
\begin{align*}
y_i &= f(\bm{X}_i) + \epsilon_i \\
f(\bm{X}_i) &= T(\bm{X}_i, \bm{\nu}, \bm{\gamma}) = \sum_{j = 1}^J I_{\bm{X}_i \in R_j} \gamma_j \ ,
\end{align*}
where the outcome variable $y_i$ for observation $i$ is some function $f$ of the vector of $p$ predictors $\bm{X}_i$.  A decision tree is most easily visualized and interpreted as a tree diagram (Figure \ref{fig:tree_mat}), but it can also be represented by the function $T(\bm{X}_i, \bm{\nu}, \bm{\gamma})$. The $J$ terminal nodes of the decision tree are represented by regions $R_j$. The vector $\bm{\nu}$ contains the split points and predictors that define regions $R_j$, and $\bm{\gamma}$ is a vector of length $J$ containing the means of the $j = 1, \hdots, J$ terminal nodes. We use the indicator function $I_{\bm{X}_i \in R_j}$ to denote that observation $i$ is within a given region, that is, $I_{\bm{X}_i \in R_j} = 1$ if subject $i$ is in terminal node $R_j$. Continuing the education example for a tree with a single split point and $J = 4$ terminal nodes, the predictor $x_{GPA}$ might have split point $\nu_1 = 3.2$, and predictor $x_{HW}$ has split point $\nu_2 = \nu_3 = .8$.\footnote{These split points are chosen for simplicity. It is rare to find that subsequent splits in each region are on the same variable and have the same cutpoint, especially when the number of predictors is large.} Four regions (or terminal nodes) $R_1, \hdots, R_4$ are defined by the split points $\{x_{GPA} < 3.2, x_{HW} < .8\}$, $\{x_{GPA} < 3.2, x_{HW} \geq .8\}$,  $\{x_{GPA} \geq 3.2, x_{HW} < .8\}$, and $\{x_{GPA} \geq 3.2, x_{HW} \geq .8\}$. The observations falling into each node are then $I_{x_{GPA} < 3.2, x_{HW} < .8}$ and so forth. The predictions of the tree are simply the mean math performance in each node. For the first node, $\hat{\gamma}_1 = \frac{1}{n_1}\sum_{i \in x_{GPA} < 3.2,  x_{HW} < .8} y_i$, where $n_1$  is the number of observations in the first node. 

In general, there is no maximum likelihood or deterministic procedure for estimating the optimal partitions of predictors into regions \citep{hyafil_constructing_1976}. Instead, a data-driven procedure called recursive partitioning is used. In the first step, the predictor with the largest main effect is selected (e.g. GPA), and the sample is split into two more homogenous regions based on a split point on that predictor (e.g. GPA $< 3.2$). The procedure continues (recursively) by selecting a new splitting variable and split point for each terminal node (e.g. homework completion rate < 80\%), and partitioning within each of the nodes. This recursive partitioning continues until a stopping criterion is reached. Common stopping criteria are (a) reaching a minimum number of observations within a node; and/or (b) carrying out a maximum number of splits. After a tree is fully grown, it can also be pruned to avoid overfitting \citep{hastie_elements_2013}. Decision trees are readily interpretable in terms of tree-diagrams, where the predictors and split points of the predictors are shown as branches (Figure \ref{fig:tree_mat}). For further details see e.g. \cite{breiman_classification_1984}, or  \cite{strobl_introduction_2009}.

Research in decision trees has further improved their performance, interpretability, and applicability to psychological data. Recursive partitioning can be improved by using statistical tests as criteria for splits and controlling Type-I error \citep{hothorn_unbiased_2006, strobl_bias_2007}. A very popular extension to decision trees is to estimate parametric models within each node. In general, this is called model-based recursive partitioning \citep{zeileis_model-based_2008}. Predictors and split points are chosen to select models with maximally different parameter estimates within each node. Examples include generalized linear models \citep{zeileis_model-based_2008}, item-response models \citep{de_boeck_irtrees:_2012}, structural equation models \citep{brandmaier_structural_2013, brandmaier_theory-guided_2016}, and generalized mixed-effect models \citep{fokkema_detecting_2015}. Applications of model-based recursive partitioning in psychology include data-driven detection of differential item functioning \citep{de_boeck_irtrees:_2012}, measurement differences over time or across groups \citep{brandmaier_structural_2013}, and estimating different treatment effects \citep{fokkema_detecting_2015}. In general, these approaches are appropriate when the general structure of the interrelations between variables is known and can be expressed in a model, and partitioning only serves to find group specific parameter estimates. If such a model is not known a priori, or the distributional assumptions of the model do not hold within each node, using boosted decision trees provides a useful and elegant alternative.

\subsection{Boosted Decision Trees}

In general, boosting is a method for estimating complex models with many parameters stagewise by {\em gradient descent} \citep{friedman_greedy_2001, friedman_additive_2000, buhlmann_boosting_2003, friedman_stochastic_2002, buhlmann_boosting_2007, elith_working_2008, hofner_model-based_2014}. Stagewise estimation proceeds by updating the model one parameter at a time without changing the previous estimates. To estimate a model stagewise by gradient descent, parameters are chosen that minimize the first derivative of the loss function (the gradient) at each step. In boosted decision trees, gradient descent is used to estimate an additive model (or ensemble) of decision trees  \citep{friedman_greedy_2001, friedman_additive_2000, friedman_stochastic_2002}. With a continuous outcome and squared error loss $L(y, \hat{y}) = \frac{1}{n} \sum_{i=1}^n (y - \hat{y})^2$, the gradient is the vector of residuals. Thus individual trees are simply fit to the residuals of the previous trees. The additive model of trees an be represented as: 
\begin{align}
y_i &= f(\bm{X}_i) + \epsilon_i \label{f_mod} \\
f(\bm{X}_i) &= \sum_{m=1}^M T_m(\bm{X}_i, \bm{\nu}_m, \bm{\gamma}_m)  \label{tree_mod},
\end{align}
where the complex, nonlinear function $f(\bm{X}_i)$ is approximated by $M$ trees with unique regions and predictions.

Estimation of an additive model of decision trees by gradient descent is necessary because there is no maximum likelihood or deterministic procedure for estimating model parameters \citep{friedman_greedy_2001}. Further, an optimal step size is unknown because derivatives cannot be taken with respect to model parameters, so that both the number of trees and the step size become {\em meta-parameters} that need to be chosen. It has been shown that stagewise gradient descent can be improved by fitting trees to a subsample of the observations at each iteration \citep{friedman_stochastic_2002}. This diminishes overfitting caused by outlying observations while speeding up computation time. An algorithm for estimating an additive model of trees by stochastic gradient descent is shown as Algorithm 1.

\begin{algorithm}
Algorithm 1: Boosted Decision Trees with Squared error Loss \citep{friedman_greedy_2001}
\begin{algorithmic}
\State $r_{i, 0} = y_i - \bar{y}$
\For{ $m = 1, \hdots, M$ steps (trees)}
 \State Let $s(i)$ be a subsample of $i \in \{1, \hdots, N\}$
 \State $(\hat{\bm{\nu}}, \hat{\bm{\gamma}})_m = \min_{\bm{\nu}, \bm{\gamma}} \sum_{s(i)} \Big( r_{s(i), m-1} - T(\bm{X}_{s(i)}, \bm{\nu}, \bm{\gamma}) \Big)^2$ 
  \State $r_{i, m} = r_{i, m-1} - \hat{T}_m(\bm{X}_i, \hat{\bm{\nu}}_m, \hat{\bm{\gamma}}_m) \lambda $  
\EndFor
\end{algorithmic}
\end{algorithm}
Like a single decision tree, the ensemble or additive model of $M$ decision trees is used to approximate an unknown function of the predictors. The ensemble of trees results in improved prediction performance compared to single trees while sacrificing the ability to interpret trees as a set of decision rules. For more discussion, see \cite{miller_finding_2016}. For computational and theoretical details, see \citep{friedman_greedy_2001, friedman_stochastic_2002}.
 
Several R, Python, and C++ implementations exist for fitting models of boosted decision trees \citep{pedregosa_scikit-learn:_2011, hothorn_mboost:_2016, hickey_gbm:_2016, chen_xgboost:_2016}. In R, one of the most popular is the package {\em gbm} \citep{hickey_gbm:_2016} which allows users to fit boosted decision tree ensembles based on different loss functions for continuous, multi-nomial, and survival outcomes. The R package {\em mvtboost} \citep{miller_finding_2016} extends gbm to continuous multivariate outcomes. In this paper, we again build on the gbm package which contains computationally efficient procedures for fitting individual decision trees.

\subsubsection{Tuning meta-parameters to avoid under- and overfitting}

A critical aspect of model fitting is selecting meta-parameters (number of trees, step size, and tree-depth) to avoid under- or overfitting the data. The model can easily overfit with a large number of trees, or can underfit if a tiny step size is specified and the maximum number of trees is too small. A minimally sufficient strategy is to choose a maximum number of trees and tree-depth corresponding to the available computation time, and then choose both the step size and the number of trees that minimizes prediction error. 

It is important to tune the maximum depth of the trees or the minimum number of observations in each node because it governs the complexity of the interactions that can be approximated by each tree. For example, a tree-depth of one (a single split) will only capture main effects, while a tree-depth of two can capture two-way interactions, etc. However, a tree-depth of two is not guaranteed to capture any particular two-way interactions because split points are chosen by selecting predictors with the largest main effects first (within each partition). Two-way and higher order interactions are typically approximated by multiple trees. In practice, choosing a tree-depth of five or ten is common. However, it is critical to tune tree-depth along with step-size and the number of trees jointly to maximize predictive accuracy and variable selection performance of the model.

The meta-parameters (number of trees, step size, and tree-depth) can be chosen by minimizing test-error or cross-validation error. The test-error is the error of model predictions on the subset of data not used for training the model. It is an accurate estimate of the prediction error, but is not efficient with small samples. Most commonly, the prediction error is estimated via $K$-fold cross validation. The $K$-fold cross-validation error is computed by splitting the observations into $k=1, \hdots, K$ sets or {\em folds}. The model is then trained $K$ times on the observations that are not in fold $k$, and the prediction error is computed for the observations within fold $k$. An estimate of the prediction error is obtained by averaging over all $K$ folds. Cross-validation is usually the best approach in practice because it uses all available data while protecting against overfitting and estimating the $K$ models can easily be carried out in parallel.  Importantly, the number of trees should never be chosen to minimize \textit{training} error (the prediction error for the subset of data used for training the model) because training error will always decrease with each additional tree.

As a practical note, the default values of the number of trees, step-size, and tree-depth provided in many implementations are primarily useful only for checking that the data format is suitable for analysis. Additionally, the commonly recommended strategy of choosing the number of trees given one step size and one tree-depth (e.g. 5 or 10) is rarely optimal. For best results, we recommend tuning the number of trees over many combinations of tree-depth, step size, and even the minimum number of observations in a node by grid search. We provide a function \texttt{gbm\_grid} for extensive tuning by grid search in the mvtboost package.

\subsubsection{Interpretation}

The final model is often interpreted by ranking the {\em relative influence} of predictors and visualizing their effects using {\em partial dependence plots}. Predictors can be ranked using relative influence, which is the contribution of each predictor to the model relative to the other predictors. For continuous outcomes, the relative influence is defined as the reductions in the sum of squared error of the outcome $(SSE)$ attributable to splits on predictors over all $M$ trees. The influence score is usually expressed as a percent of the total reductions in $SSE$ due to splits on all predictors. It has been shown that predictor selection by relative influence provides good variable selection performance that balances true and false positive rates when the predictors are on the same scale \citep{miller_finding_2016}. However, the naive relative influence is known to favor predictors with many categories \citep{strobl_bias_2007}. A conditional importance framework is promising alternative, which uses p-values testing the association of an outcome with a proposed split as the split criterion \citep{hothorn_unbiased_2006, strobl_conditional_2008}. This approach has not yet been implemented or evaluated for boosted decision trees.

Another way that the additive model of trees can be interpreted is through the use of partial dependence plots. A partial dependence plot illustrates the model-implied effect of an individual predictor, averaging over the effects of other predictors. See \cite{friedman_multiple_2003} for further details and \cite{goldstein_peeking_2015} for independent conditional expectation plots. Though ensembles of decision trees are models of interactions (each split is conditional or depends on the previous split), it is difficult to visualize sets of many interacting predictors. Current methods for detecting sets of interacting variables from the model are limited to two-way interactions and are only heuristics \citep{elith_working_2008, miller_finding_2016}. The performance of these heuristics for detecting two-way interactions is unknown.

\section{Linear mixed effect model}

Next, we briefly introduce the linear mixed effect model. Let there be $i = 1, \hdots, g$ groups with $j = 1, \hdots, n_i$ observations in each group. Let $\bm{y}_{i}$  be a vector of length $n_i$, containing observations from group $i$, and let $N = \sum_i n_i$ be the total number of samples. A general mixed effect model for $\bm{y}_i$ is given by \citep{bates_fitting_2015,  pinheiro_mixed-effects_2006}:

\begin{align}
\label{lme}
\bm{y}_i &= \bm{X}_{i} \bm{\beta} +  \bm{Z}_i \bm{b}_i + \bm{\epsilon}_i \\
\bm{b}_i & \overset{iid}{\sim} MN(\bm{0}, \bm{\Psi}) \nonumber \\
\bm{\epsilon}_i &\overset{iid}{\sim} MN(\bm{0}, \sigma^2 \bm{I}_{n_i})  \nonumber
\end{align}
Where $\bm{y}_i$ is an $n_i \times 1$ vector of responses for group $i$, $\bm{X}_i$ is a $n_i \times p$ matrix of predictors, and $\bm{\beta}$ is a $p \times 1$ vector of fixed effects for $p$ covariates. $\bm{Z}_i$ is the $n_i \times q$ design matrix for the random effects, with the $q \times 1$ vector of weights $\bm{b}_i$ for the random effects for each group $i = 1, \hdots, g$. The $\bm{\epsilon}_i$ is the $n_i \times 1$ vector of errors. The $q \times q$ matrix $\bm{\Psi}$ is the variance covariance matrix of the random effects. In matrix form, the model can be expressed as

\begin{equation}
\bm{y} = \bm{X \beta} + \bm{Z b} + \bm{\epsilon},
\end{equation}
where $\bm{y}$ is an $N \times 1$ vector of responses from the entire sample, $\bm{X}$ is an $N \times p$ matrix of predictors (including intercept), and $\bm{\beta}$ is a $p \times 1$ vector of fixed effects. The matrix $\bm{Z}$ is an $N \times (gq)$ matrix of random effects with $(gq) \times 1$ vector of weights $\bm{b}$.

\subsection{Estimation}

When the variance components $\bm{\Psi}$ are known, the model $(\ref{lme})$ can be estimated as the solution to Henderson's mixed model equations \citep{henderson_best_1975}. 

\begin{equation}
\begin{bmatrix}
\bm{X}'\bm{X} & \bm{X}' \bm{Z} \\
\bm{Z}'\bm{X} & \bm{Z}'\bm{Z} + \bm{\Psi} \\
\end{bmatrix}
\begin{bmatrix}
\bm{\beta} \\ \bm{b}
\end{bmatrix}
= 
\begin{bmatrix}
\bm{X}'\bm{y} \\
\bm{Z}'\bm{y}
\end{bmatrix}
\end{equation}
It can be shown that these estimates are the best linear unbiased predictions of the random effects \citep{henderson_best_1975, searle_matrix_1997}. When the variance components must also be estimated, the model is usually estimated by restricted maximum likelihood. See \cite{bates_fitting_2015} for estimation details specific to lme4.

The estimated random effects from the mixed effect model are shrunken toward 0 when group sizes are small or when the within group variance is large. To illustrate, consider the simple random intercept model $\bm{y}_i = \mu + \alpha_{i} + \bm{e}_i$ where $\bm{y}_i \sim N(\alpha_{i}, \sigma^2)$ and  $\alpha_{i} \sim N(\mu, \sigma^2_{\alpha})$. The variance component $\sigma^2_{\alpha}$ is the between group variance, and $\sigma^2$ is the within group variance. Then the mixed effect model estimates $\hat{\alpha}_{i}^*$ of the group-specific parameters $\alpha_{i}$ are $\bar{y}_i - \mu$, which are shrunk toward the grand mean $\mu$ by a factor $\omega$ \citep{searle_matrix_1997}:
\begin{equation}
\hat{\alpha}_{i}^* =  \frac{\sigma^2_{\alpha}}{\sigma^2_{\alpha} + \frac{\sigma^2}{n_i}} (\bar{y}_{i} - \mu ) =  \omega (\bar{y}_{i} - \mu ) \label{shrinkage} 
\end{equation}
As $\omega \rightarrow 1$, the between group variance $\sigma^2_{\alpha}$ is much larger than the within group variance $\sigma^2$, and the mixed effect model estimates $\hat{\alpha}_{i}^*$ approach the deviations of the group means from the grand mean. As $\omega \rightarrow 0$, the within group variance $\sigma^2$ is much larger than the between group group variance $\sigma^2_{\alpha}$, and the mixed effect model estimates $\hat{\alpha}_{i}^* \rightarrow 0$. In finite samples with unknown variances, the additional $n_i$ term is incorporated to account for the precision of the estimate of the within group variance $\sigma^2$. Small group sizes result in lower precision, which in turn results in lower $\omega$ and more shrinkage of $\hat{\alpha}^*_i$ toward 0.
 
\section{metboost: mixed effect tree boosting }

Next, we describe a procedure for extending boosted decision trees to allow predictors to have random effects. This procedure is called metboost, and combines decision trees with mixed effect models by constraining the split points to be the same across groups (e.g. GPA $< 3.2$ across all schools) but allowing the terminal node means of the tree to vary by group (GPA $< 3.2$ has different implications for math performance for different schools). Nonlinear effects are approximated by the terminal node means of each tree. Adding a grouping variable allows group specific terminal node means to approximate group specific nonlinear effects.

In addition to approximating group specific nonlinear effects, treating grouping variables as random rather than fixed addresses common research questions in psychology. It also increases the efficiency of the predictions from each tree when group sizes are small. Additionally, metboost is helpful when the number of predictors is large and when it is unclear which predictor effects should be allowed to vary by group. Simply allowing all predictors to vary by group is computationally intensive, and it is also challenging to estimate such a large number of variance components unless the group sizes are very large. To address this problem, metboost iteratively selects predictors with random effects by gradient descent.

The procedure works by fitting an additive model of trees, where the split points of each tree are the same for each group, but the terminal node means differ. This is accomplished at each iteration in the following way. First a decision tree is fit to the data ignoring the grouping and variable, and this tree is then represented as a design matrix. Each column of the design matrix is an indicator variable denoting terminal node membership. The effects of the indicator variables are then allowed to vary by group in a mixed effect model. Finally, the fitted values from this mixed effects decision tree are used to update the predictions at each iteration. 

\begin{figure}
\includegraphics[scale=1]{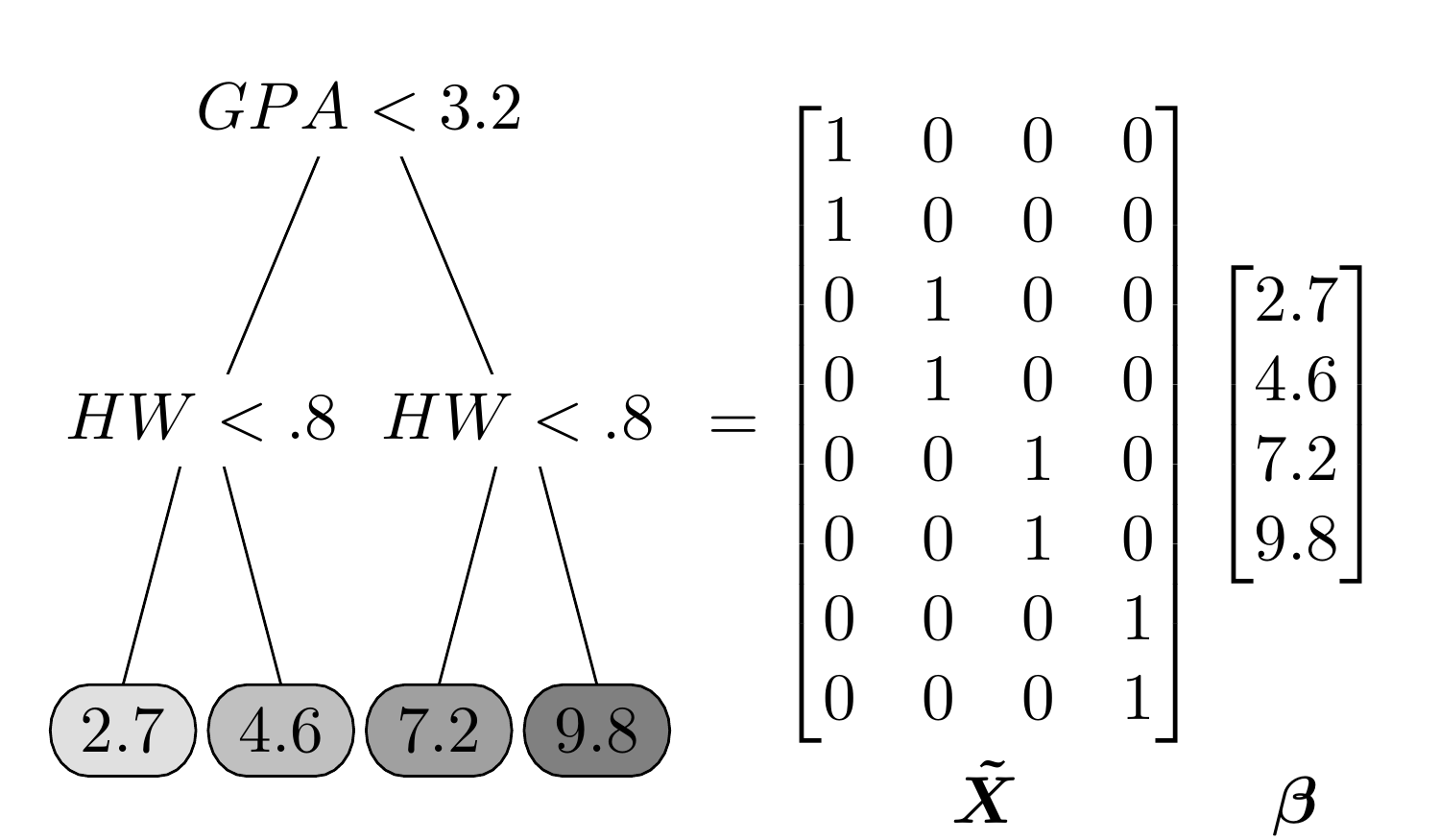}
\caption{Representing a decision tree with four terminal nodes and two observations per node as a design matrix $\bm{\tilde{X}}$ multiplied by a weight vector $\bm{\beta}$. Each column of $\bm{\tilde{X}}$ is an indicator that assigns an observation to a terminal node. The weight vector $\bm{\beta}$ contains the terminal node means.}
\label{fig:tree_mat}
\end{figure}

\begin{figure}
\includegraphics[scale=1]{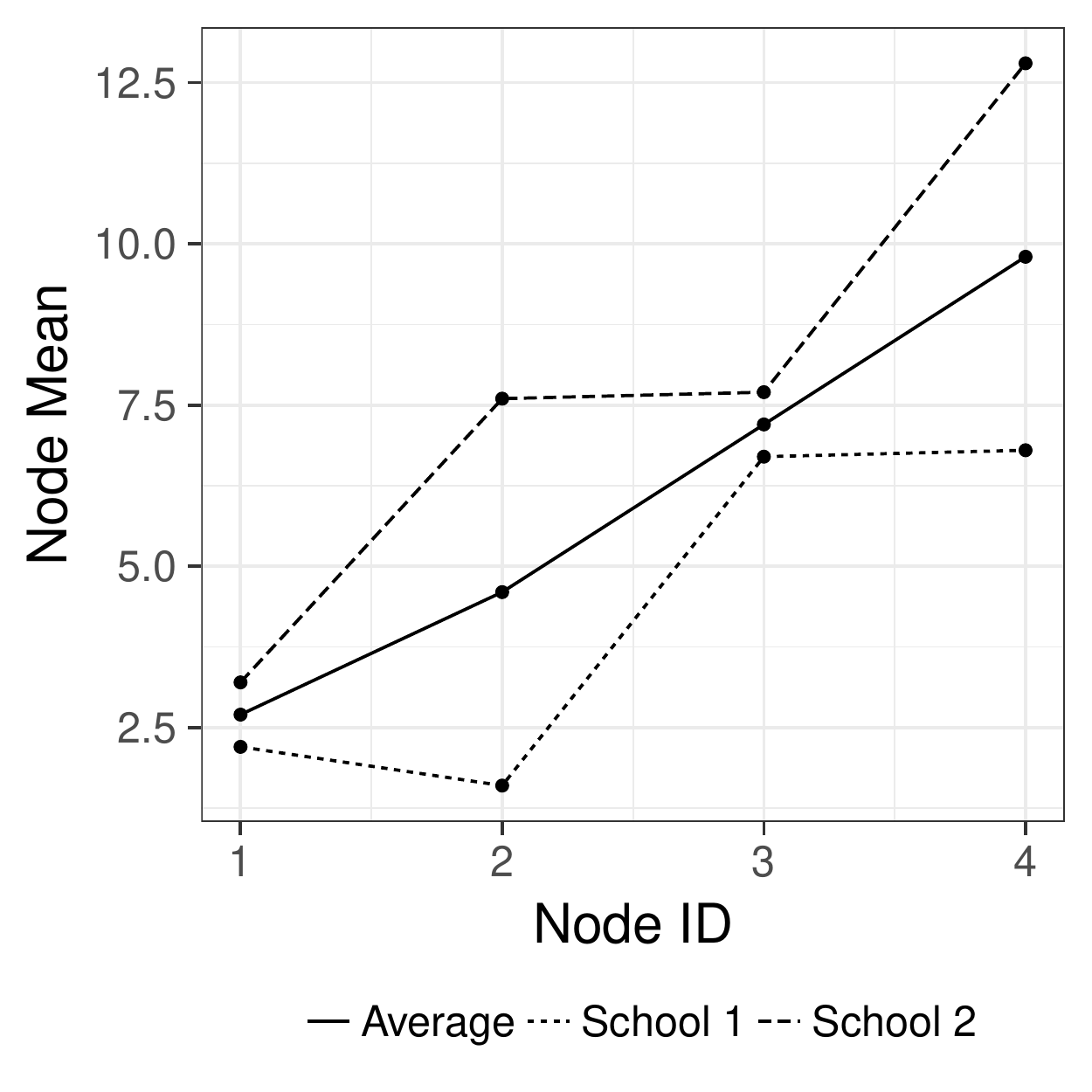}
\caption{Predictions for decision tree with four terminal nodes. The decision tree predictions averaged across schools is shown as a solid line. School specific predictions are shown as dotted lines.}
\label{fig:tree_mat_rand}
\end{figure}

To illustrate this approach, consider again the example of predicting math performance on a standardized test from 8 high school seniors from $i = 1, 2$ schools. A single tree is fit to this data and has 4 terminal nodes (Figure \ref{fig:tree_mat}). For example, the first split is GPA $< 3.2$. Within each node, a second split is chosen for homework completion rate < 80\%. These splits can be represented as a design matrix $\tilde{\bm{X}}$ with $k = 4$ columns and 8 rows (Figure \ref{fig:tree_mat}). The full tree can be represented by $\tilde{\bm{X}} \bm{\beta}$, where $\bm{\beta}$ contains the mean math performance for each node (Figure \ref{fig:tree_mat}). The tree captures nonlinear effects of predictors that are the same across schools by allowing the patterns of means in the terminal nodes to take any functional form. To allow the effect of GPA and homework completion rate to differ by school, the indicators (columns) in this design matrix are allowed to vary by group. In terms of the formula syntax from lme4, such a model is represented by:

\begin{lstlisting}[language=R]
MATH ~ X1 + X2 + X3 + X4 + (X1 + X2 + X3 + X4 | SCHOOL)
\end{lstlisting}
In this model, the random effects are the {\em group specific} terminal node means (see e.g. Figure \ref{fig:tree_mat_rand}).  In terms of the linear mixed effect model matrices \citep{bates_fitting_2015}, the random effects design matrix $\bm{\tilde{Z}}_i$ for schools one and two in Figure \ref{fig:tree_mat_rand} would be as in (\ref{eq:zi})
 \begin{align}
\bm{\tilde{Z}}_1 = \begin{bmatrix}
1 & 0 & 0 & 0 \\
1 & 0 & 0 & 0\\
0 & 1 & 0 & 0\\
0 & 1 & 0 & 0\\
\end{bmatrix}
\bm{\tilde{Z}}_2 = \begin{bmatrix}
0 & 0 & 1 & 0 \\
0 & 0 & 1 & 0\\
0 & 0 & 0 & 1\\
0 & 0 & 0 & 1\\
\end{bmatrix}
\label{eq:zi},
\end{align}
where the random effects $\bm{b}_i$ were assigned as $\bm{b}_1 = [-.3, -.5, 0, 0]', \bm{b}_2 = [0, 0, .3, .5]$.  These random effects allow the splits GPA $< 3.2$ and homework completion rate < 80\% to have different implications for math performance in each school. The limitation is that only a small number of predictor effects are accounted for in any single tree. In this example, only the effects of GPA and homework completion are accounted for. Thus to allow for the model to account for nonlinear group specific effects of many predictors, many of these trees can be fit by gradient descent (i.e. boosting).

The desirable properties of metboost are highlighted in this example. First, group specific nonlinear effects are detected with each tree. Group specific nonlinear effects for all predictors are accounted for in the ensemble of trees. Second, the design matrix $\bm{\tilde{X}}$ is guaranteed to be fully observed because it is a matrix of indicators assigning observations to terminal nodes. Observations with missing values on a splitting variable are assigned to nodes based on a split on a surrogate. Finally, the model is faster to estimate than a mixed effect model containing random effects for $p$ predictors because the number of random effects is constrained to be the number of terminal nodes $k$. In our example, the original sample could have had as many as 50 or 100 predictors, but the design matrix for any given tree is restricted to 4 columns, the number of terminal nodes. This constraint results in  a significant improvement in computation time.

More generally, the model can be understood as follows:
\begin{align}
\bm{y}_{i} &= f(\bm{X}_i) + g_i(\bm{X}_i) + \bm{\epsilon}_{i} \label{f_gi} \\
f(\bm{X}_i) &= \sum_{m=1}^M T_m(\bm{X}_i, \bm{\nu}_m, \bm{\gamma}_m) \nonumber \\
&= \tilde{\bm{X}}_{i, m} \bm{\beta}_m  \nonumber \\
g_i(\bm{X}_i) &= \sum_{m=1}^M T_m(\bm{X}_i, \bm{\nu}_m, \bm{\gamma}_{m, i}) \label{gi_tree} \\
&= \sum_{m = 1}^M \tilde{\bm{Z}}_{i, m} \bm{b}_{i, m}  \label{gi_lmer},
\end{align}
where $f$ is a function of predictors constrained to be the same across groups, while $g_i$ are unknown functions of the predictors unique to each group. The functions $g_i$ are approximated by allowing the means of the terminal nodes in tree $m$ to vary by group, $\bm{\gamma}_{m, i}$ (\ref{gi_tree}). 

Allowing the means of the terminal nodes to vary by group is accomplished by representing tree $m$ as a special mixed effect model with design matrices $\tilde{\bm{X}}_{i, m}, \tilde{\bm{Z}}_{i, m}$ (\ref{gi_lmer}). 
A tree can be viewed as a categorical variable mapping individuals to terminal nodes. This categorical variable is equivalently represented by a design matrix of indicator variables (Figure \ref{fig:tree_mat}). We denote this mapping of the partitions in a tree to a design matrix of indicators by the operator $D\big(T(\bm{X}_i, \bm{\nu}, \bm{\gamma}) \big)$. The resulting design matrix is denoted $\tilde{\bm{X}_i}$. For a single decision tree with $k$ terminal regions, $\tilde{\bm{X}}_i$ has $k$ columns of indicator variables, and $n_i$ rows (Figure \ref{fig:tree_mat}).

The matrix $\tilde{\bm{Z}}_i$ allows the terminal node means of the tree to vary by group. It is constructed by multiplying each column in $\tilde{\bm{X}}_i$ by vectors of group indicators \citep{bates_fitting_2015}. If $\bm{J}_i$ is the $n_i \times k$ matrix of indicator variables for group membership, the design matrix for the random effects $\tilde{\bm{Z}}_i$ is the $n_i \times k$ matrix obtained by element wise multiplication of each column in $\tilde{\bm{X}}_i$ by each column in $\bm{J}_i$. Thus, $\tilde{\bm{X}}_{i, m}$ is $n_i \times k$ design matrix implied by the decision tree $m$, and $\tilde{\bm{Z}}_{i, m}$ is the $n_i \times k$ design matrix allowing the means of the nodes $\bm{\gamma}_m$ to vary by group. The weights $\bm{\beta}_m$ are a $k \times 1$ vector of means for each terminal node implied by the tree $m$, and the $\bm{b}_{i, m}$ is a $q \times 1$ vector of group specific deviations from the terminal node mean. A summary of the matrices in this model appears in Table 1.

For example, the model matrices in (\ref{f_gi}) from the data in Figures \ref{fig:tree_mat} and \ref{fig:tree_mat_rand} is given by:

\begin{align}
\tilde{\bm{X}} \bm{\beta} + \tilde{\bm{Z}} \bm{b} &= 
\begin{bmatrix}
\tilde{\bm{X}}_1 \\
\tilde{\bm{X}}_2 \\
\end{bmatrix}
\begin{bmatrix}
\bm{\beta}  \\
\end{bmatrix} +
\begin{bmatrix}
\tilde{\bm{Z}}_1 &  \bm{0} \\
\bm{0} & \tilde{\bm{Z}}_2 \\
\end{bmatrix} 
\begin{bmatrix}
\bm{b}_1 \\ \bm{b}_2
\end{bmatrix}\\
&=
\begin{bmatrix}
1 & 0 & 0 & 0 \\
1 & 0 & 0 & 0\\
0 & 1 & 0 & 0\\
0 & 1 & 0 & 0\\
0 & 0 & 1 & 0\\
0 & 0 & 1 & 0\\
0 & 0 & 0 & 1\\
0 & 0 & 0 & 1\\
\end{bmatrix}
\begin{bmatrix}
2.7 \\ 4.6 \\ 7.2 \\ 9.8 \\
\end{bmatrix} +
\begin{bmatrix}
1 & 0 & 0 & 0 &    &    &     &    \\
1 & 0 & 0 & 0 &    &    &     &    \\
0 & 1 & 0 & 0 &    &    &     &   \\
0 & 1 & 0 & 0 &    &    &     &  \\
   &    &    &    & 0  & 0 &  1 & 0  \\
   &    &    &    & 0  & 0 &  1 & 0  \\
   &    &    &    & 0  & 0 &  0 & 1 \\
   &    &    &    & 0  & 0 &  0 & 1  \\
\end{bmatrix} 
\begin{bmatrix}
-.5 \\ -3 \\ 0 \\ 0 \\ 0 \\ 0 \\ .3 \\ .5
\end{bmatrix}
\label{tilde_mats}
\end{align}
The columns in $\bm{\tilde{X}}$ are comprised of $k = 4$ columns of indicators assigning individuals to terminal nodes in a tree using the first node as a reference. The terminal node means of the tree are $\bm{\beta} = \bm{\gamma}$.  These represent, for instance, the average math performance on a test for all students with GPA $ < 2.3$ and homework completion rate $< 80\%$. The matrix $\bm{\tilde{Z}}$ is partitioned by group with blocks $\bm{\tilde{Z}}_i$. The matrix $\tilde{\bm{Z}}$ is given by a multiplication of each column in $\bm{\tilde{X}}$ by the group indicators $\bm{j}'_1 = [1, 1, 1, 1, 0, 0, 0, 0]$, $\bm{j}'_2 = [0, 0, 0, 0, 1, 1, 1, 1]$. The vector of random effects parameters $\bm{b}$ allows terminal node means parameterized by $\bm{\beta}$ to vary by group membership. In our example, this allows the mean math performance of students with GPA $> 2.3$ to vary by school.

\begin{table}
\caption{Description of matrices in linear mixed effect models and mixed tree models.}
\begin{tabular}{l l l}
\toprule
Symbol & Dimensions & Meaning \\
\midrule
$g$ &  & Number of groups in the grouping variable \\ 
$n_i$ & & Number of observations within group $i$ \\
$N = \sum_i n_i$ &  & Total number of observations \\
$p$ & & Number of predictors including intercept  \\
$k$ & & Number of terminal nodes \\
$q$ & & Number of random effects \\
%$\bm{y}_i, \bm{y}$ & $n_i \times 1, N \times 1$ vector of outcomes \\
$\bm{X}_i, \bm{X}$ & $n_i \times p, N \times p$ & Design matrix of fixed effects plus intercept \\
$\bm{J}_i, \bm{J}$ & $n_i \times g, N \times g$ & Matrix of indicator variables denoting group membership \\
$\bm{Z}_i, \bm{Z} $ & $n_i \times q, N \times (gq)$ & Design matrix of random effects \\
$\bm{\tilde{X}}_i, \bm{\tilde{X}}$ & $n_i \times k, N \times k$ & Design matrix of fixed tree effects \\
$\bm{\tilde{Z}}_i, \bm{\tilde{Z}}$ & $n_i \times k, N \times (gk)$ & Design matrix of random tree effects \\
\bottomrule
\end{tabular}
\end{table}

\subsubsection{Estimation by gradient descent}

This model (\ref{f_gi}) can be estimated iteratively by gradient descent. In step $m$ a single tree $T_m(\bm{X}_i, \bm{\nu}_m, \bm{\gamma}_m)$ is fit, and its design matrix computed. Next, the means within nodes $\bm{\gamma}$, are allowed to vary by group by fitting a mixed effect model with tree $m$ as the design matrix to the residuals at step $m$ as follows:

\begin{equation} 
	\bm{r}_{i, m} = \tilde{\bm{X}}_{i, m} \bm{\beta}_m + \tilde{\bm{Z}}_{i, m} \bm{b}_{i, m} +  \bm{\epsilon}_{i, m} \label{res_lmer}
\end{equation}
Where $\bm{r}_{i, m}$ is the $n_i \times 1$ vector of residuals at iteration $m$, $\tilde{\bm{X}}_{i, m} = D\big(T_m(\bm{X}_i, \bm{\nu}_m, \bm{\gamma}_m) \big)$ is the $n_i \times k$ tree design matrix, $\bm{\beta}_m$ is the $k \times 1$ vector of node means, and $\bm{b}_{i, m}$ is the $k \times 1$ vector of deviations within each group from the terminal node mean. The estimates of the parameters $\bm{\beta}_m$ and $\bm{b}_{i, m}$ can be obtained using general purpose mixed effect modeling software packages, such as the R package lme4 \citep{bates_fitting_2015}. The estimates for $f(\bm{X}_i)$ and $g_i(\bm{Z}_i)$ at iteration $m$ are given by:

\begin{align}
\hat{f}_m(\bm{X}_i) &=  T_m(\bm{X}_i, \hat{\bm{\nu}}_m, \hat{\bm{\gamma}}_m) \nonumber \\
&=  \tilde{\bm{X}}_i \hat{\bm{\beta}}_m \label{fix_update} \\
\hat{g}_{i, m}(\bm{Z}_i)  &= T_m(\tilde{\bm{X}}_{i, m}, \hat{\bm{\nu}}_m, \hat{\bm{\gamma}}_{i, m}) \nonumber \\ 
&= \tilde{\bm{Z}}_{i, m} \hat{\bm{b}}_{i, m}  \label{random_update}
\end{align}
Estimating both $\hat{f}, \hat{g}_i$ using squared error loss by gradient descent proceeds iteratively by fitting a new tree to the residuals of the previous fit as follows:

\begin{equation}
\label{update}
\bm{r}_{i, m} = \bm{r}_{i, m - 1} - \lambda \Big( \hat{f}_m(\bm{X}_i) + \hat{g}_{i, m}(\bm{Z}_i) \Big) 
\end{equation}
Where $\bm{r}_{i, m}$ is the $n_i \times 1$ vector of residuals at iteration $m$, and $\lambda$ is a fixed step size at some small value, e.g. .01. In the first iteration, $\bm{r}_{i, 0} = \bm{y}_i - \bar{y}$. Like with univariate boosting with squared error loss (Algorithm 1), subsampling from each group at each iteration can improve performance. The resulting stochastic gradient descent algorithm is summarized in Algorithm 2.
\begin{algorithm}
Algorithm 2: Stochastic gradient descent for boosted decision trees with hierarchically clustered data
\begin{algorithmic}

\State $\bm{r}_{0} = \bm{y} - \bar{y}$
\For{ $m = 1, \hdots, M$ steps (trees)}
 \State Let $s(N) = sub\_sample(1, \hdots, N)$ be a subsample of all observations
 \State $\hat{\bm{\nu}}_m, \hat{\bm{\gamma}}_m = \min_{\bm{\nu}, \bm{\gamma}} \sum_{s(N)} \Big( \bm{r}_{m-1} - T(\bm{X}, \bm{\nu}, \bm{\gamma}) \Big)^2$ \Comment{(Algorithm 1)}
 \State $\bm{\tilde{X}}_m = D\big(T(\bm{X}, \hat{\bm{\nu}}_m, \hat{\bm{\gamma}}_m)\big)$ \Comment{Mapping tree to design matrix}
 \State $\hat{\bm{\beta}}_m, \hat{\bm{b}}_{m} = \min_{\bm{\beta}, \bm{b}} \sum_{s(N)}  L\Big(\bm{r}_{m}, \tilde{\bm{X}}_{m} \bm{\beta} + \tilde{\bm{Z}}_{m} \bm{b} \Big) $ \Comment{(\ref{res_lmer})}
 \State $\hat{f}_m(\bm{X}) = \tilde{\bm{X}} \hat{\bm{\beta}}$,  \hspace{.2in} $\hat{g}_{m}(\bm{Z}) =  \tilde{\bm{Z}}_{m} \hat{\bm{b}}_{m}$  \Comment{ (\ref{fix_update}, \ref{random_update})}
 \State $\bm{r}_{m} = \bm{r}_{m - 1} - \lambda \Big( \hat{f}_m(\bm{X}_i) + \hat{g}_{m}(\bm{Z}) \Big) $   \Comment{(\ref{update})}
\EndFor
\end{algorithmic}
\end{algorithm}

\subsection{Interpretation}

The two methods of interpreting the resulting model are through ranking predictors according to their influence on the model and by plotting the model implied effects of individual predictors. For a continuous outcome variable with squared error loss, the influence of predictor $j = 1, \hdots, p$ is the sum of the reductions in $SSE$ due to splits on this predictor, summed over all trees in the model \citep{friedman_greedy_2001}. The influence scores for all $p$ predictors are often scaled to sum to 100, reflecting the percent in total reductions in $SSE$ due to each predictor. Defining the influence in this way captures the fixed effects of predictors, because split points in each tree are the same in each group. To visualize potentially nonlinear effects of predictors,  the model predictions $\hat{y}$ can be plotted against one or two predictors of choice at a time. These plots show the the marginal effect of each predictor on the model predictions. If the grouping variable is of direct interest, separate plots can be created showing how the marginal effect of a predictor varies by group. 

\subsection{Implementation}

An implementation of metboost is provided in the R package {\em mvtboost}  \citep{miller_finding_2016}.\footnote{The metboost method from {\em mvtboost} is not yet on CRAN, but will be before publication. For review purposes, a development version can be installed from \texttt{github.com/patr1ckm/mvtboost}.} This implementation provides easy meta-parameter tuning by cross validation in parallel, and is built on the newest version of the R package {\em gbm} \citep{hickey_gbm:_2016}. Fitting the model with cross-validated meta-parameter tuning in parallel can be done as follows:

\begin{lstlisting}[language=R]
 metboost(y = y, X = X, id = "id", 
 	n.trees = 2500,
	interaction.depth = c(3, 5, 8),
	shrinkage = c(.005, .01, .025), 
	bag.fraction = .5,
	cv.folds = 3,
	subset = 1:500, 
	mc.cores = 12)
\end{lstlisting}

In this example \texttt{y} is a vector of outcomes with no missing data, \texttt{X} is a matrix of predictors optionally containing missing values, and \texttt{id} is name of the grouping variable in \texttt{X}.  Following the specification of the data, the other key meta-parameters are listed, including the number of trees, the tree-depth, and shrinkage values. These meta-parameters are the same as in gbm and similar to other boosting algorithms. Setting \texttt{cv.folds} > 1 allows cross validation over the grid implied by the vectors of values given to \texttt{n.trees}, \texttt{interaction.depth}, and \texttt{shrinkage} (step-size). Setting \texttt{mc.cores} carries out the cross-validation in parallel for a given number of cores. The quantities $\hat{\bm{y}}$, $\bm{X}\hat{\bm{\beta}}$, and $\bm{Z} \hat{\bm{u}}$ for the best number of trees and best set of meta-parameters with the lowest cross-validation error are returned.

In addition, to the metboost function, an \texttt{influence} function is provided for computing the influence of important variables. A \texttt{plot} function is provided that plots the predicted values against predictors to help detect the presence of nonlinear effects. A \texttt{predict} function is also provided that computes predicted values at a given number of trees. However, it is faster and more memory efficient to simply compute predicted values along with training the model by indicating which observations are used for training and testing by the \texttt{subset} argument. 

\begin{figure}
\includegraphics[scale=.75]{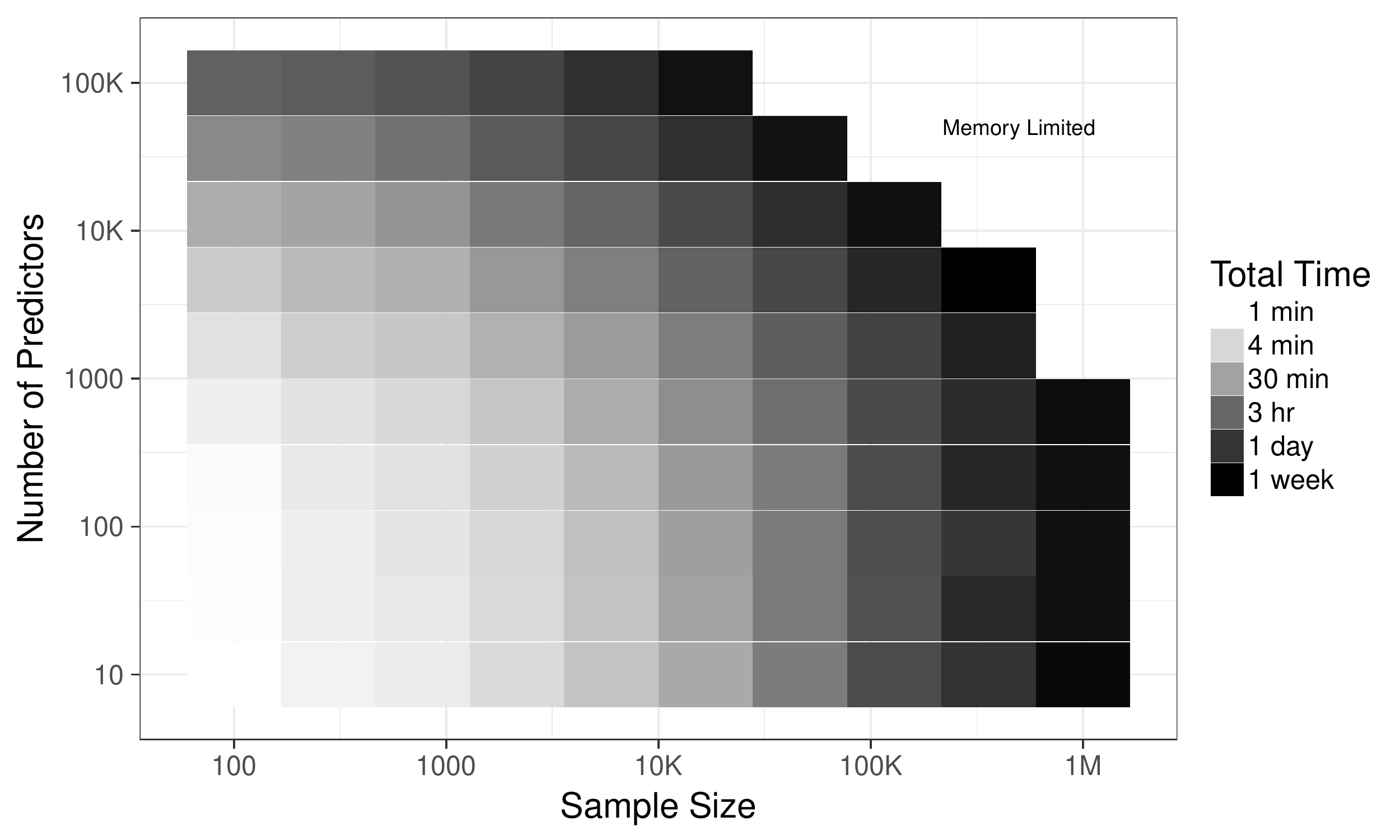}
\caption{Computation time for metboost as a function of sample size and the number of predictors for 1000 trees.  These timings were carried out on Quad 16 core 2.4 GHz AMD Opteron processors with up to 128 GB of RAM. Trees were fit with 8 threads. Timings do not include cross-validation over a grid of meta-parameters. For small to moderate data sets, analyses can be carried out in a few hours. When both the number of samples and the number of predictors is large, computation time can take a week or longer.}
\label{fig:timing}
\end{figure}

Estimates of computation time for metboost are shown in Figure \ref{fig:timing} for a range of sample sizes and number of predictors for 1000 trees. The group size was always 50. These timings were carried out on Quad 16 core 2.4 GHz AMD Opteron processors with up to 128 GB of RAM. For small data sets ($n < 1000$ and $p < 100$), metboost can be run in 1-4 minutes. For  moderate data sets, ($n < 10,000$ and $p < 1000$), analyses can be carried out in a few hours or less. However, very large samples ($n > 100,000$) or a very large number of predictors ($p > 10,000$) can require days or weeks of computing time. For a computation time estimate with more trees, the estimates in Figure \ref{fig:timing} can be extrapolated linearly.

To illustrate the advantage of taking the hierarchical structure of data into account, we compare metboost to gbm in a simulation study next. Following the simulation study, we use the method to predict math ability using data from the Educational Longitudinal Study 2002.

\section{Simulation Study: Prediction Error and Variable Selection Performance}
In this simulation study, the variable selection and prediction performance of metboost and gbm were compared while varying the following factors when generating the data: the number of predictors, the type of effect (linear only or nonlinear), the number of predictors with random effects, average group size, and the $ICC$ (intra-class correlation coefficient). In the next sections, details are provided on how the artificial data sets were generated, the overall simulation design, and how the methods were compared.

\subsection{Simulation Design}
\subsubsection{Data generation}
Artificial data sets were simulated under the following mixed effect model (\ref{lme}, repeated):
\begin{align}
\bm{y} &= \bm{X} \bm{\beta} + \bm{Z} \bm{b} + \bm{e}  \nonumber \\
\bm{b} &\overset{iid}{\sim}  MN(0, \bm{I}_q) \nonumber \\
\bm{e} &\overset{iid}{\sim}  MN(0, \sigma^2 \bm{I}_N) \nonumber
\label{dg}
\end{align}
The predictors were independent, normally distributed variables with mean 0 and variance 1. Of the total number of predictors $P$, a smaller number $p$ was chosen to have nonzero fixed effects. Of these $p$ predictors, a subset $q$ were chosen to have effects that varied by group. With the variance of the intercepts and slopes fixed to 1, the error variance $\sigma^2$ was chosen so that random effects had a chosen ICC as follows:
\begin{align}
ICC &= \frac{1}{1 + \sigma^2} \\
\sigma^2 &= \frac{1 - ICC}{ICC}
\label{icc}
\end{align}
The fixed effect $\beta_j$ for predictor $j = 1, \hdots, p$ was chosen to control for a given $\tilde{R}^2$ as follows:
\begin{align}
\tilde{R}^2 &= \frac{Var(\bm{X}\bm{\beta})}{Var(\bm{X}\bm{\beta}) + Var(\bm{Z} \bm{b}) + \sigma^2} \\
\beta_j &= \frac{\tilde{R}^2 (\bm{Z}'\bm{Z} + \sigma^2)}{(1 - \tilde{R}^2)p} 
\label{r2}
\end{align}
To generate nonlinear fixed effects, the design matrix $\bm{X}^*$ was used in place of $\bm{X}$, and was created by transforming predictor $j = 1, \hdots, p$ as $\bm{x}_j^* = f(\bm{x}_j)$. The function was chosen randomly from the following set: $f_0(x) = x$, $f_1(x) = x^2$, $f_2(x) = \sqrt{|x|}$, $f_3(x) = \cos(\pi x)$, $f_4(x) = |\sin(\pi x)|$. Functions $f_1, f_2$ were chosen so that the predictor $\bm{x}^*_j$ would have a main effect of $0$. Functions $f_3, f_4$ were chosen to simulate periodic behavior, while constraining predictors to have a main effect of $0$. Function $f_0$ was included to incorporate predictors with truly linear or predictors with effects that can be linearly approximated (e.g. $e^x$, $x^3$). 

\begin{figure}
\includegraphics[scale=1]{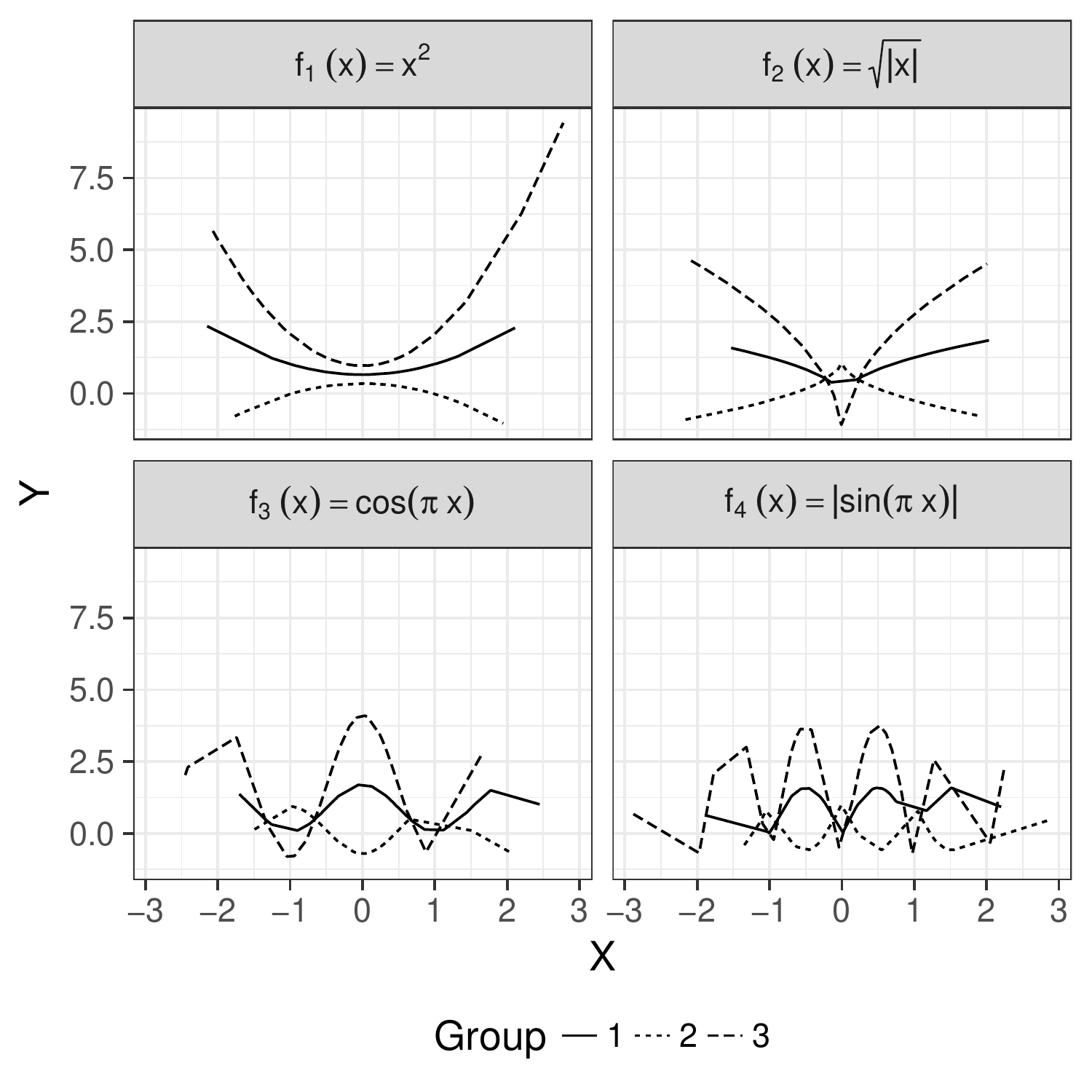}
\caption{Example group specific nonlinear effects generated for a single covariate and three groups using four nonlinear functions. }
\label{fig:sim_data}
\end{figure}

To generate group specific linear or nonlinear effects for $q$ predictors, the matrix $\bm{Z}$ was generated as 
$$\bm{Z} = \bm{X}^*_{1, \hdots, q} * \bm{J}$$
Where $\bm{J}$ is the indicator matrix denoting group membership (i.e. column $\bm{j}_i = 1$ for observations in group $i$). $\bm{X}^*_{1, \hdots, q}$ is the subset of $q$ columns in $\bm{X}^*$ (nonlinear) or $\bm{X}$ (linear) that have group specific effects. The operator $*$ denotes the column-wise product. Finally, group specific linear or nonlinear functions were generated by multiplying $\bm{Z b}$, with unique weights $\bm{b} \overset{iid}{\sim} MN(0, \bm{I}_q)$. Examples of the simulated group specific nonlinear effects are shown in Figure \ref{fig:sim_data}.

\subsubsection{Simulation Factors}
The levels of the five factors used in the study are summarized in Table \ref{tab:sim_design}. As in the applied example, both metboost and gbm were fit to each generated data set. The best number of trees and shrinkage values were chosen by 3-fold cross-validation in the training set over a grid of meta-parameter values. The values of the meta-parameters used for gbm and metboost are summarized in Table \ref{tab:sim_metap}. The candidate values for each method were chosen based on pilot simulations, and to ensure reasonable computation time. For each method, the relative influence scores for each predictor and predicted values for observations in the test set are based on the best number of trees was chosen from the set of meta-parameters that achieved the lowest cross-validation error. 
\begin{table}[ht]
\setlength\tabcolsep{2pt}
\begin{center}
\caption{Simulation Design.}
\label{tab:sim_design}

\begin{tabular}{p{8cm} p{4cm}}
\toprule
\multicolumn{1}{c}{Factor}  & \multicolumn{1}{c}{Levels} \\
\midrule
Number of predictors 		 &  5, 25, 50, 100, 250 \\
Effect type   				 & Linear, Nonlinear \\
Predictors with random effects	 &  5, 10, 25, 50 \\
Average group size 			 & 4, 5, 8, 10, 20, 40 \\
ICC 						 & .3, .5, .8 \\
Total effect size  			 & .5 \\
Fraction of predictors  nonzero effects& 25 \\
N 						 &1,000 \\

\bottomrule
\end{tabular}
\end{center}
\bigskip
\small\textit{Note:} Five varying factors were setup in a factorial experiment, resulting in 720 simulation conditions. Replicated data sets were generated from (\ref{dg}). The error variance was chosen to control the ICC (\ref{icc}), and the fixed effect size for predictors with nonzero effects was chosen to control the total effect size (\ref{r2}). The number of replications was 50.
\end{table}

\begin{table}[ht]
\setlength\tabcolsep{2pt}
\begin{center}
\caption{Simulation meta-parameter grids for gbm and metboost.} 
\label{tab:sim_metap}
\begin{tabular}{p{9cm}p{4cm}p{3cm}}
\toprule 
\multicolumn{1}{c}{Meta-parameter}   & \multicolumn{1}{l}{gbm}  &  \multicolumn{1}{l}{metboost} \\
\midrule
Number of trees & 10,000  & 2,500 \\
Shrinkage &  .005, .01, .05, .1  & .01, .025, .05, .1 \\
Tree-Depth &  5, 10, 25  & 3, 5 \\
Minimum number of observations in node &  $\tilde{n}_i$,  20   & 20 \\
\bottomrule
\end{tabular}
\end{center}
\bigskip
\small\textit{Note:}
gbm was run with 24 unique combinations of meta-parameters, while metboost was run over 8 unique combinations. metboost requires a much smaller tree-depth, and fewer number of trees than gbm. In addition to shrinkage and tree-depth, gbm was also tuned over the minimum number of observations in each terminal node. These values were the default (20) and the average group size, $\tilde{n}_i$.
\end{table}

The methods were compared in terms of their mean squared prediction error on a test set of equal size as the training set ($n_{train} = n_{test} = 1000$). The methods were also compared on how well they correctly selected important variables, which is referred to as their {\em variable selection performance.} Variable selection performance is quantified by the area under the ROC curve (AUC) for variable selection, which is based on the relative influence scores from each predictor calculated for each model. The influence scores for each predictor are computed from the trees in both metboost and gbm as usual. However, the relative influence scores for gbm were calculated without including the influence of the grouping variable. Without this adjustment, the influence of many of the other predictors from gbm would be close to 0. This is has been previously observed in practice \citep{strobl_bias_2007}.

The AUC ranges from .5 to 1, with .5 being chance variable selection and 1 being perfect variable selection. It can be computed as follows. Given a vector of influence scores for $p$ predictors and a vector of known truly significant predictors, predictors can be selected with influence greater than some threshold $\tau$. This results in a $2 \times 2$ confusion matrix containing counts of all correct (true positives, true negatives) and incorrect decisions (false positives, false negatives) in the cells. For any given $\tau$, the true and false positive rates for variable selection can be computed. A ROC curve is then used to plot the true positive rate against the false positive rate for all possible thresholds. An optimal ROC curve has an AUC of 1. In this case the true positive rate is equal to 1 regardless of the false positive rate as thresholds for selection become more lenient. An AUC of .5 corresponds to chance variable selection. In this case the false positive rate increases linearly with the true positive rate as thresholds become more lenient.

\subsection{Simulation Results}

\begin{figure}
\includegraphics[scale=1]{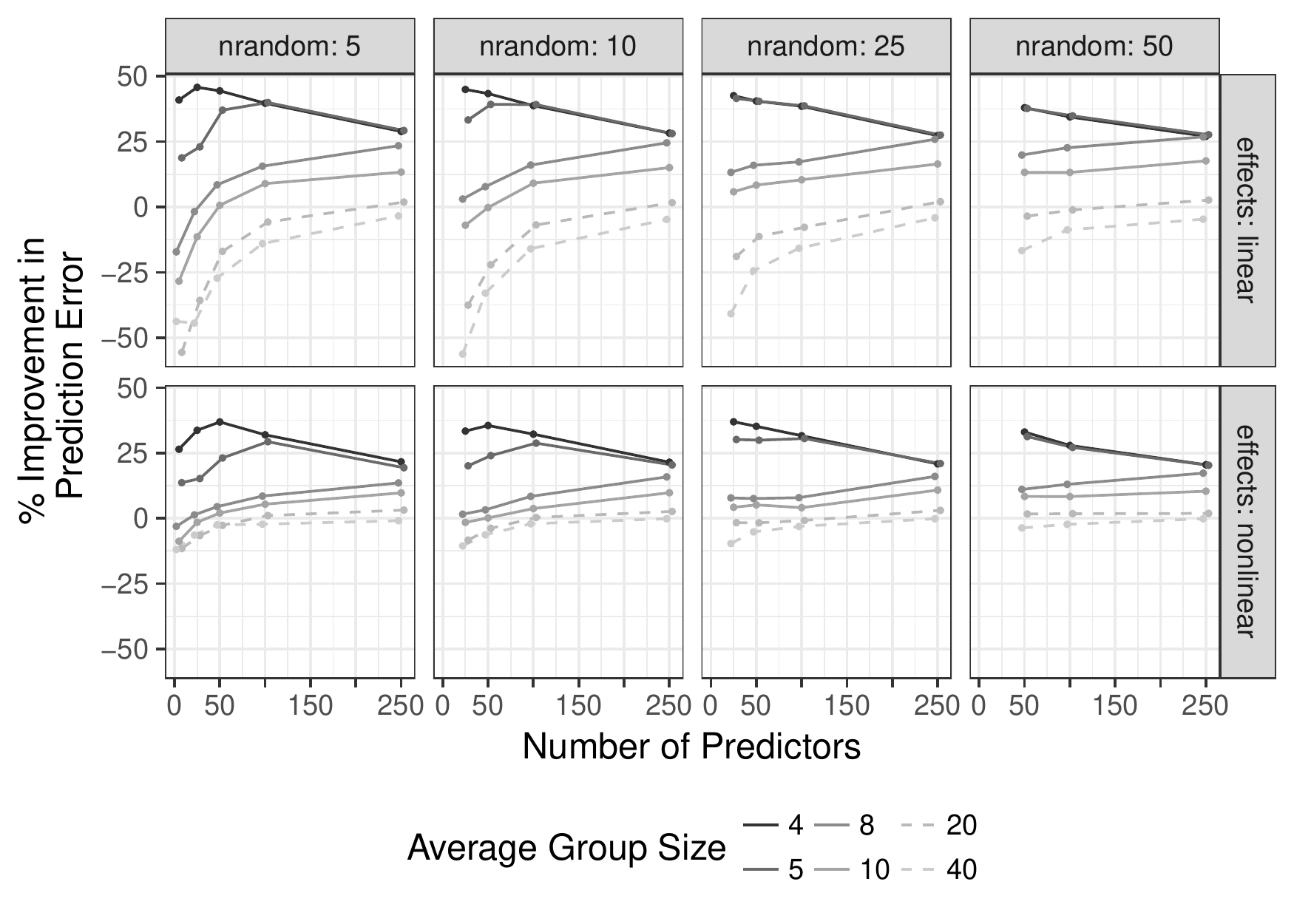}
\caption{Percent improvement in prediction error of metboost compared to gbm (y-axis), plotted as a function of the number of predictors (x-axis), the number of random effects (columns), the kind of effect (rows), and average group size. Conditions with scores above 0 are favorable for metboost, and are indicated with solid lines. Results are shown for $n = 1000$, $ICC = .5$, and with predictors explaining $\approx 50\%$ of the variance in the outcome. Results show that gbm performs better than metboost only with large group sizes, a small number of predictors, and explicitly linear effects (top left panels), otherwise metboost performs better. metboost can perform as much as 35\% better when group sizes are small, and almost always performs better when group specific effects are nonlinear.}
\label{fig:sim_mspe}
\end{figure}

\begin{figure}
\includegraphics[scale=1]{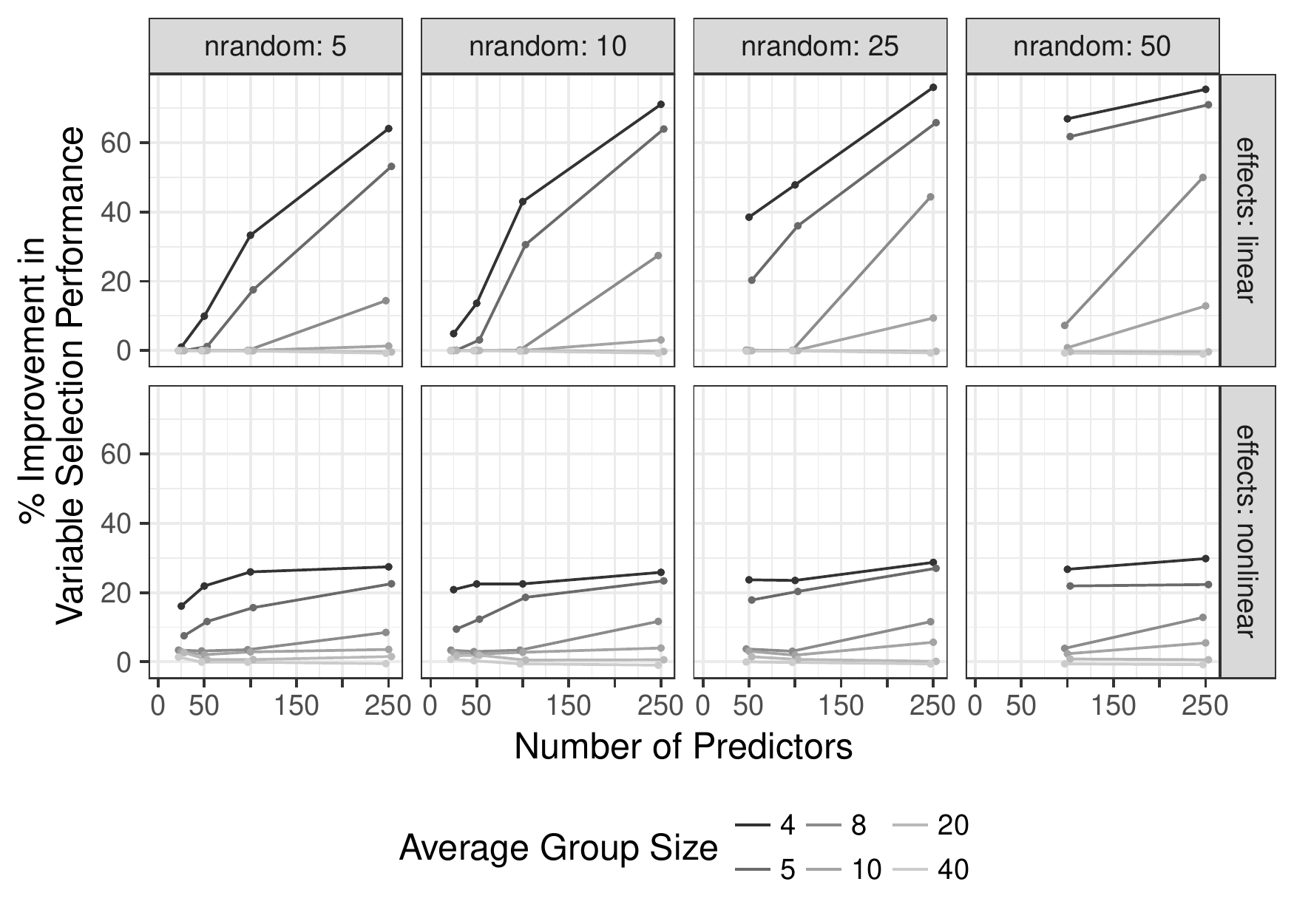}
\caption{Percent improvement in variable selection performance of metboost compared to gbm (y-axis) plotted as a function of the number of predictors (x-axis), the number of random effects (columns), the kind of effect (rows), and average group size. Conditions with scores above 0 are favorable for metboost, and are indicated with solid lines. Results are shown for $n = 1000$, $ICC = .5$, and with predictors explaining $\approx 50\%$ of the variance in the outcome. Results show that metboost has uniformly better variable selection performance across this range of conditions.}
\label{fig:sim_vsp}
\end{figure}

The percent change in prediction and variable selection performance for metboost relative to gbm is shown in Figures \ref{fig:sim_mspe} and \ref{fig:sim_vsp} at ICC $= .5$. Performance of both methods improved with larger ICCs. However, the ICC did not discriminate between the methods or interact with the other factors. Simulation results show that metboost has uniformly higher variable selection performance over gbm across all conditions, achieving up to 70\% better variable selection performance when group sizes are small and effects are linear (Figure \ref{fig:sim_vsp}, top row). When effects are nonlinear and group sizes are small, metboost achieves up to a 20\% improvement in variable selection performance (Figure \ref{fig:sim_vsp}, bottom row). When group sizes were large, metboost had 1-5\% improvement over gbm when the number of predictors increased (Figure \ref{fig:sim_vsp}).

The results also show that metboost can achieve improvements in prediction performance of 10-30\% over gbm  when the average group size is small and when there are group specific nonlinear effects (Figure \ref{fig:sim_mspe}, bottom row). When effects were exactly linear, metboost could achieve 25\% improvements or more when group sizes were small Figure \ref{fig:sim_mspe}, top row). However, when group sizes were large, effects were linear, and the number of predictors is between 5-50, gbm outperforms metboost by 30-50\% (Figure \ref{fig:sim_mspe}, top row). This is larger than expected, but diminishes to a 0-10\% improvement over metboost as the number of predictors increase. We expect that this result will not be critical in practice because the random and fixed effects of many predictors are unlikely to be exactly linear. Further, if a small number of predictors were known to have exactly linear effects, the true model is the linear mixed effect model and no exploratory regression analysis is necessary. 

Next, we compare metboost and gbm to a large data set from education. This data set illustrates the motivating factors for using metboost for exploratory regression analysis: predictors may have group specific effects, the effects could be nonlinear, many predictors have missing data, and there are a large number of predictors and samples.

\section{Application of metboost to Education Longitudinal Study 2002}

A natural application of metboost is for exploratory analyses of data sets in education, where the effects of student level predictors can often be expected to differ between schools. To illustrate this application, a data set on performance on a standardized math test was obtained from the Education Longitudinal Study from the base year 2002 (ELS), from the National Center for Education Statistics \citep{ingels_education_2007}. This nationally representative sample included responses from 17,591 students, 7,135 teachers, and 743 principals from 751 schools \citep{ingels_education_2007}. For this analysis, the student is the primary unit of analysis. Of the original 17,591 students, 15,240 had an observed math standardized score. Students from 751 unique schools were included in the dataset, and the number of students in each school ranged from 2-50 with median 20. The dependent variable of interest in this analysis is the performance of students on a math standardized test included in the ELS. This was the ability estimate from a 2-parameter item response model, which was re-normalized to a z-score with 5.33\% of the scores imputed. Details of the imputation procedure are provided in \cite{ingels_education_2007}. 

Of the 4937 variables collected in the survey (including original items, item aggregates, and survey weights), 109 publicly available items and aggregates of items were selected for evaluation of their utility in predicting the math ability estimates. Groups of variables are summarized in Table \ref{tab:edu_data}. Seven new aggregate scores were created as sum or factor scores from multiple items. This was done for interpretability, to reduce measurement error, to decrease collinearity, to reduce the impact of missing values on narrowly worded items, and to improve computation time at the expense of some predictive accuracy. Of the new aggregate scores, 3 aggregates were simple counts of the number of math competency exams in grades 8-12, number of awards or participations in competitions, and the number of attendance problems respectively. One variable was added indicating dropout from the high school before graduation. Next, 3 factor scores were generated from their respective items: Enjoys Math, Teacher Rated Attendance, and Teacher Rated Tardiness. The items used to form these 7 aggregate scores (39 in total) were dropped from the analysis, leaving 76 predictors plus the school identifier for inclusion in the prediction models (Table \ref{tab:edu_data}).

\begin{table}[ht]
\setlength\tabcolsep{2pt}
\begin{center}
\caption{Summary of included variables from the Education Longitudinal Survey 2002. }

\begin{small}
\begin{tabular}{llll}
\toprule
\multicolumn{1}{l}{\textbf{Rater}} &  \multicolumn{1}{l}{N Vars} & \multicolumn{1}{c}{Representative Items} \\
\multicolumn{1}{l}{\hspace{.1in} Name of Group} & & \\
\midrule
\textbf{Student} 							& 51 &  \vspace{.1in} \\ 
\hspace{.1in} Math Class Format 		& 12 & problem solving in class, Class Preparation Scale \\
\hspace{.1in}  Demographics 			& 9 & sex, parent ed, SES, In IB/AP Program \\ 
\hspace{.1in}  Characteristics 			& 7 & awards, attention problems, GPA, dropout, no. of risk factors \\
\hspace{.1in} Beliefs \& Attitudes		& 7 & highest degree expected, math self efficacy, Enjoys Math \\
\hspace{.1in}  Extra-Curricular  			& 5 & no. of school activities, held job for pay \\
\hspace{.1in}  School/Homework 		& 5 & homework hours in school, hours spent on homework\\
\hspace{.1in}  Motivation 				& 2 & enjoys math, instrumental motivation  \\
\hspace{.1in}  Teacher Assessment 		& 2 & teacher quality,  teacher/student relationship \\
\hspace{.1in}  Beliefs \& Values of Friends	& 2 & friends consider grades important,  friends dropped out \\

\midrule
\textbf{Teacher} 							& 17 & \vspace{.1in} \\
\hspace{.1in} Ratings of student		& 6 & highest degree expected, student falling behind, `has disability' \\
\hspace{.1in} Education \& Experience	& 5 & total years teaching, highest graduate degree held \\ 
\hspace{.1in} Philosophy \& Motivation	& 3 & can learn to be good at math, would teach math again \\
\hspace{.1in} Characteristics			& 2 & sex, days missed \\
\hspace{.1in} Classroom				& 1 & difficulty of class \\
\midrule
\textbf{Principal}  							& 8 & \vspace{.1in} \\
\hspace{.1in} School Characteristics 		& 6 & public/private, region, \% of 10th graders with free lunch \\
\hspace{.1in} Safety \& Climate			& 2 & school safety, academic climate \\
\midrule
Total  & 76 & \\
\bottomrule
\end{tabular}
\label{tab:edu_data}
\end{small}
\end{center}
\small\textit{Note:}  Variables are grouped by rater, and then by face validity (1st column).
\end{table}

\subsection{Analysis}

The metboost algorithm was used to predict the math standardized score from the 76 predictors shown in Table \ref{tab:edu_data}. The effects of these predictors were allowed to vary by school. metboost was trained on 80\% of the sample ($n_{train} = 12,192$ observations), with 2500 trees over the following grid of meta-parameters: tree-depth = (3, 5, 8), shrinkage = (.005, .01, .025, .05, .1) for a total of 15 unique combinations of meta-parameters. The best set of meta-parameters was chosen by 3-fold cross-validation within the training set, and the best number of trees chosen to from that set to minimize cross-validation error.

We compare the results of predicting the math standardized score using metboost to the results of gbm including the school id variable as a candidate for splitting. Like metboost, gbm was tuned by 3-fold cross-validation over the following grid of meta-parameters: tree-depth = (5, 10, 25, 49),  shrinkage = (.005, .01, .025, .05, .1), minimum number of observations in each node = (10, 25, 50). The number of trees was 5000. This resulted in 60 total configurations of meta-parameters. As with metboost, the best set of meta-parameters was chosen by 3-fold cross-validation within the training set, and the best number of trees chosen to from that set to minimize cross-validation error. 

\subsection{Results}

The best set of meta-parameters for metboost was tree-depth = 3, shrinkage = .025, and the best number of trees was 2388. For gbm, the best set of meta-parameters was interaction.depth = 5, shrinkage = .025, minimum number of observations in each node = 50, and the best number of trees was 4115. The mean squared prediction error in the test set ($n_{test} = 3048$)  achieved by metboost was .40, compared to .47 for gbm, an improvement of 15\%. In metboost, school accounted for 8\% of the variance in predictions. The improvement in prediction performance of metboost in the test set is consistent with simulation results when group sizes are relatively large and nonlinear effects are present (15\%). 

\begin{figure}

\includegraphics[scale=.8]{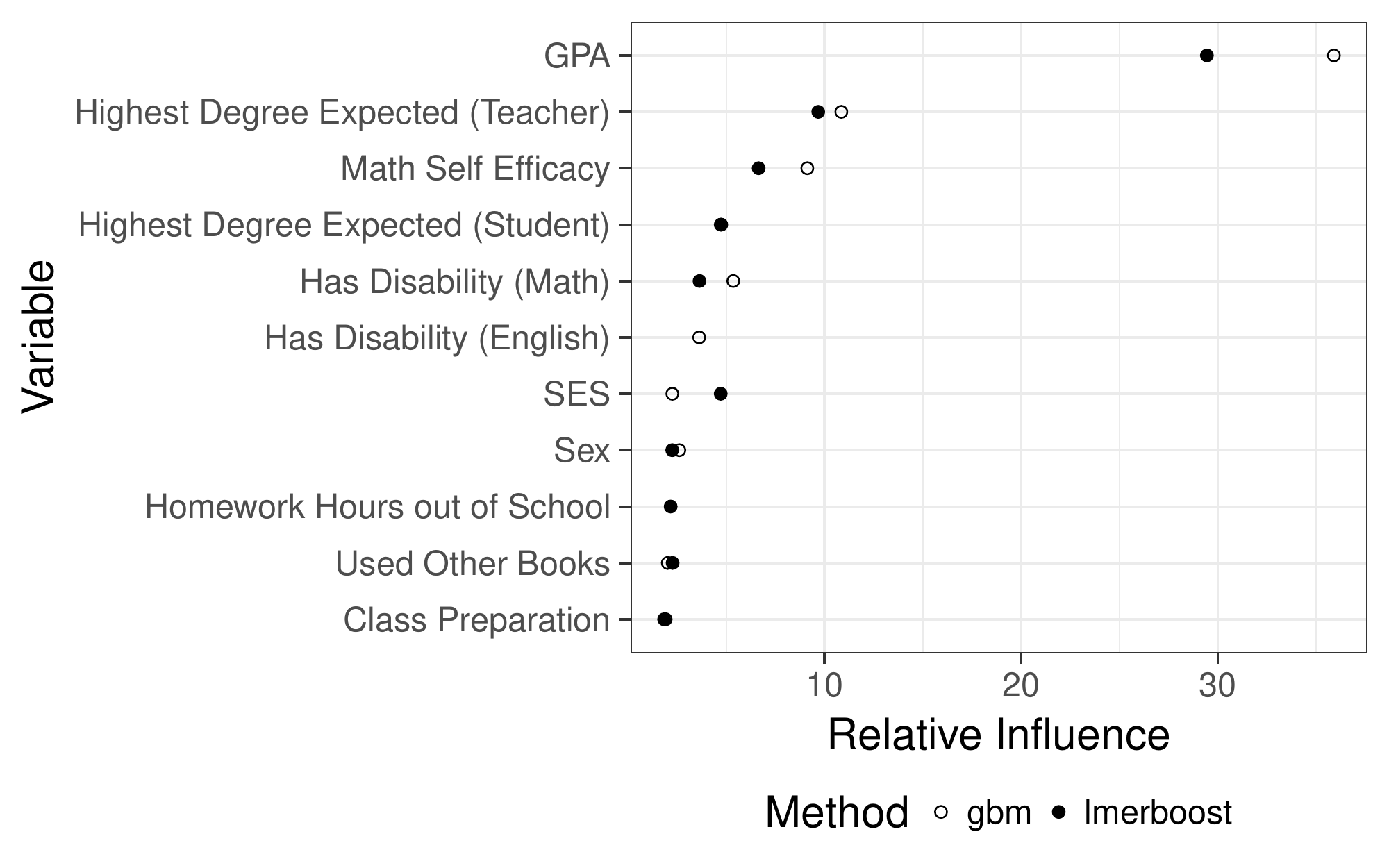}
\caption{The relative influence (x-axis) of the top ten (out of 76) predictors on the math ability estimate. The relative influence scores are compared between metboost (filled) and gbm (open). The gbm influence score was computed removing the influence of school. Results are similar in rank except that highest degree expected (student) is ranked 4th for metboost but not gbm, while `has disability' (rated by the student's english teacher) is ranked 5th for gbm but not for metboost. }
\label{fig:math_ri}
\end{figure}

The relative influence scores for the top 10 predictors from metboost and gbm are shown in Figure \ref{fig:math_ri}. For the relative influence scores to be comparable between gbm and metboost, the influence of school was set to 0 in gbm. Of the 76 predictors, the most important predictor was GPA, followed by highest degree expected (teacher) and math self efficacy. The rank correlation between the influence scores for all predictors from gbm and metboost was $\rho = .89$, $S = 7757.1$, $p < 2.2e^{-16}$. The only discrepancy in rank between the top 10 predictors was that highest degree expected (student) is ranked 4th for metboost but unranked for gbm, while `has disability' (rated by the student's english teacher) is ranked 5th for gbm but unranked for metboost. 

\begin{figure}

\includegraphics[scale=.6]{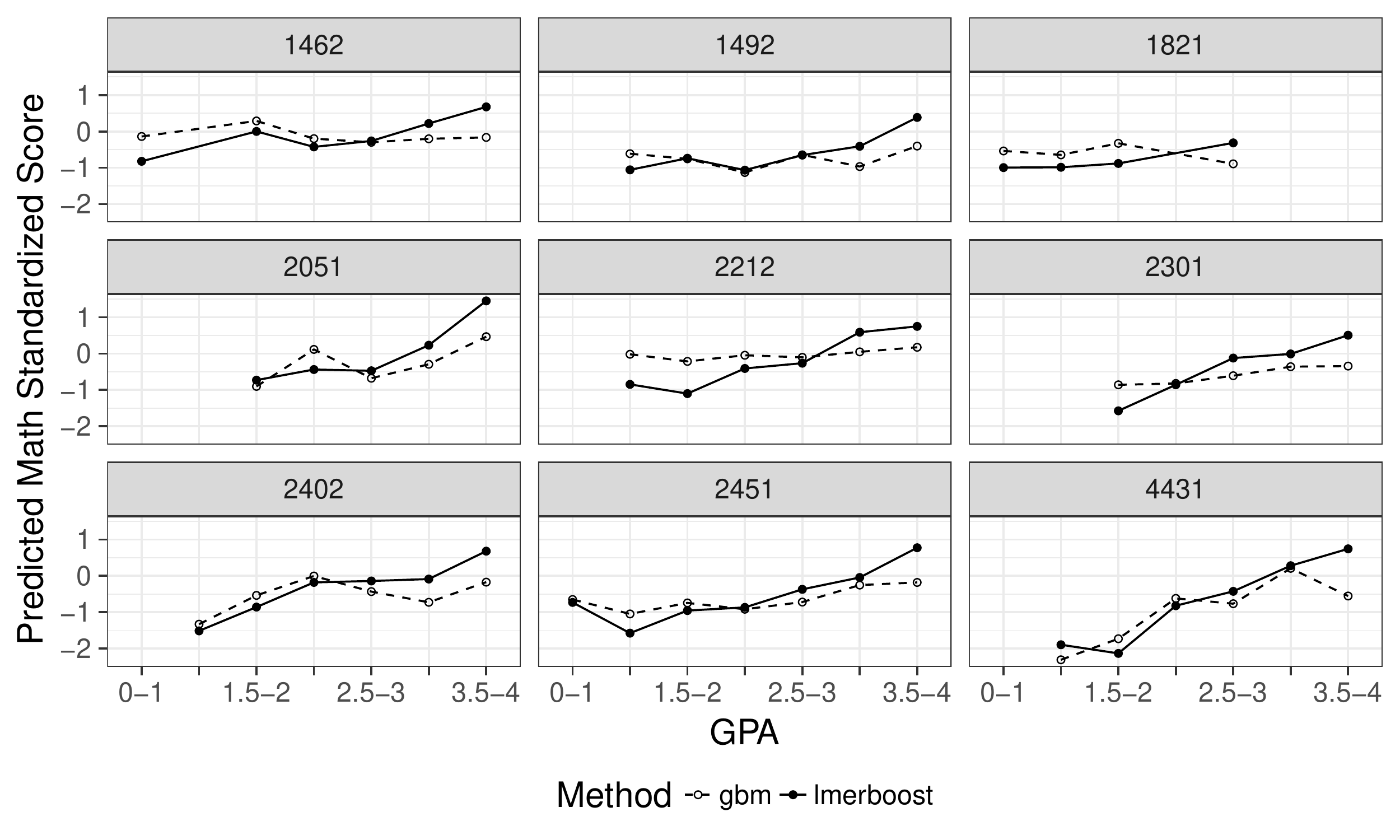}
\caption{The model implied effect of GPA on the math standardized score for nine schools. These nine schools chosen had the most discrepant predictions between gbm and metboost.}
\label{fig:math_gpa}
\end{figure}

\begin{figure}

\includegraphics[scale=.6]{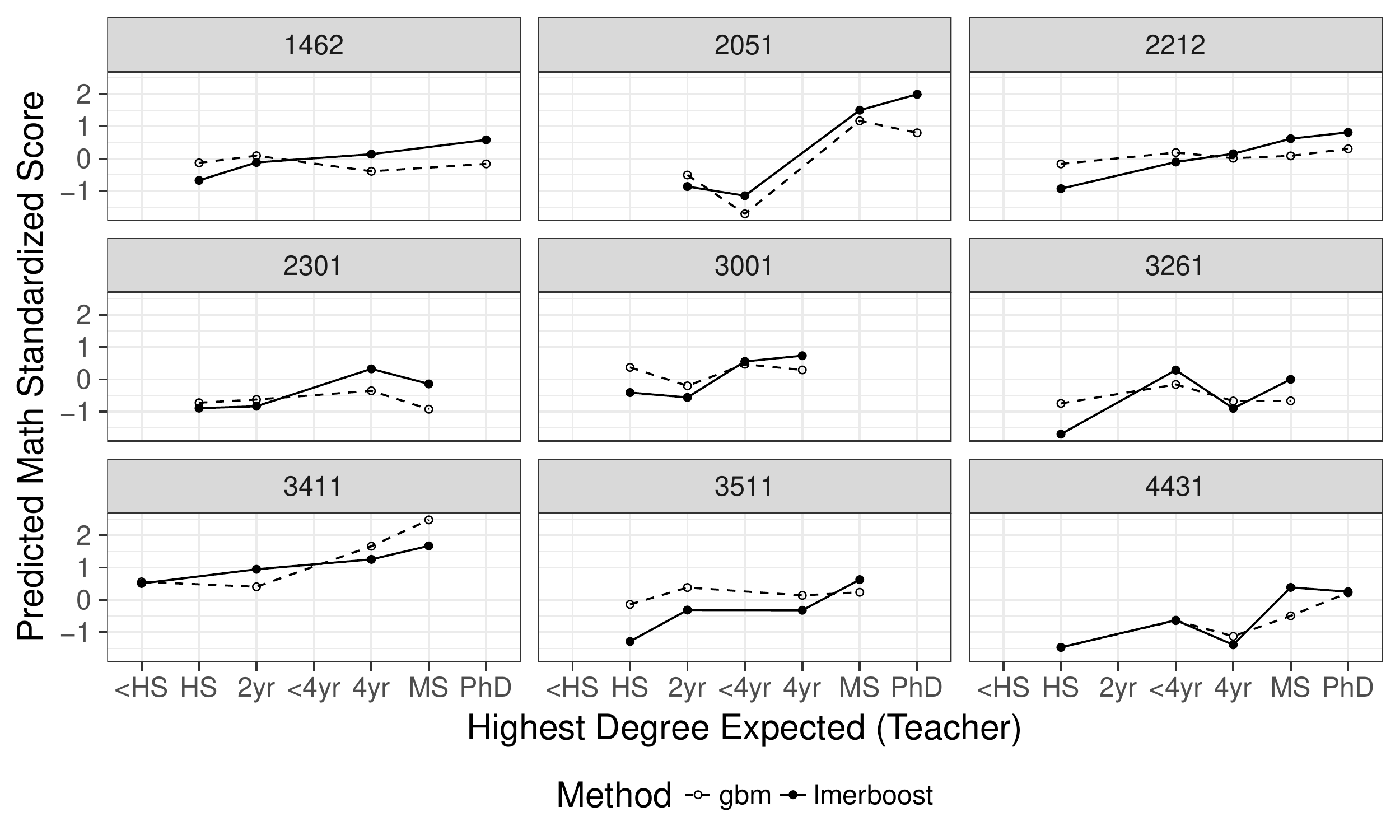}
\caption{The model implied effect of highest degree expected (teacher)  on the math standardized score for nine schools. These nine schools chosen had the most discrepant predictions between gbm and metboost.}
\label{fig:math_te}
\end{figure}

The model implied effects of GPA and highest degree expected (teacher) are shown in Figures \ref{fig:math_gpa} and \ref{fig:math_te}. These plots suggest that the effects of GPA and highest degree expected (teacher) on math performance are approximately linear. These model implied effects are shown for nine schools with the most discrepant predictions between gbm and metboost. While no consistent patterns of differences between the predictions emerged, the metboost predictions are more theoretically informative because school is treated as random. 

\section{Discussion}

Exploratory regression analysis with mixed effect models is an important but difficult problem. As the number of variables in psychological research increase (such as with a large educational study), exploratory questions becomes more important. Which predictors should be included in the model? Which predictor effects should be random or fixed? A good exploratory regression analysis should be able to identify predictors with nonlinear effects and predictor effects that vary by group, handle missing data, while remaining computationally feasible for a large number of predictors and samples. Plots and model selection can address some of these questions, but they require strong assumptions, a separate data imputation step if missing data are present, or they cannot handle a large number of predictors or samples. There are many existing parametric and semi-parametric models that can approximate group specific nonlinear effects, but almost all require a priori model specification or selection. Additionally, few existing methods are computationally feasible when the number of predictors is large and work well with missing data. 

To address this gap, we have proposed a method called metboost for exploratory regression analysis, which combines the strengths of boosted decision trees with mixed effect models. Built on boosted decision trees, metboost yields excellent prediction and variable selection performance for predictors with nonlinear effects even when observations are missing on some predictors. Group-specific nonlinear effects are included in the model by constraining the split points of each tree to be the same across groups, but allowing the terminal node means to differ. This feature allows splits (e.g. GPA $< 3.2$) to have the same cut point but different implications for the outcome (e.g. math performance for each school). These modified trees are then fit in an ensemble by gradient descent (boosting), which increases variable selection and prediction performance compared to individual trees. The only other method that achieves all the goals of exploratory regression analysis are decision tree ensembles such as boosted decision trees \citep{friedman_greedy_2001} with the grouping variable as a candidate for splitting, in which group specific effects are chosen in a data driven way.

As a demonstration, metboost was used to predict math performance on a standardized test in a large educational data set from the Educational Longitudinal Survey 2002 \citep{ingels_education_2007}. This data set had a moderate number of samples ($> 10,000$) and predictors (76), many of which contained missing values. Collectively, all 76 included predictors achieved a 40\% prediction error in a test set, which was a 15\% improvement in prediction error relative to gbm. In the context of machine learning, such large gains are difficult to achieve. Both metboost and gbm selected GPA and highest degree expected by teacher as being the most important predictors, and had consistent predictor ranks overall.  Importantly, math performance varied significantly across the schools considered in the study, with school accounting for 8\% of the variance in the predictions. 

In the more general settings considered the simulation, results showed that metboost had uniformly better variable selection performance compared to gbm across group size, number of predictors, and number of predictors with random effects (up to 70\% better). These results indicated that even though variable selection was similar in the Educational Longitudinal Survey data, metboost can more consistently select a correct set of variables to be included in a subsequent model regardless of the inclusion threshold chosen. In terms of prediction performance, metboost had up to 50\% improved prediction over gbm when group sizes were small. When group sizes were large, gbm and metboost had similar prediction error in the presence of nonlinear effects. There was a single case when gbm outperformed metboost: when the number of predictors was small (5-50), the number of predictors with group specific effects was small (5-10), group sizes were large (> 20), and effects were exactly linear. In this case, gbm had 25-50\% better prediction performance than metboost. However, in this situation, even better performance can be achieved by simply using a linear mixed effect model (the true model). In such cases, a nonparametric exploratory procedure is not necessary.

There are several theoretical limitations of the method which could be addressed in future research. First, metboost does not produce a definitive conclusion regarding which predictors should be treated as random or fixed, and only indicates a subset of candidate predictors that could be considered random. We note, however, that identifying such a subset is difficult using existing methods when the number of predictors is large. A second limitation is that many of the mixed models fit in the ensemble do not converge. The effect of convergence on variable selection or prediction error could be explored, as well as different solutions. A third limitation is in interpretation. For example, predictors can only be ranked by their influence score. However, this is essentially true of all exploratory procedures (even model selection), and is advantageous compared to interpreting p-values after model selection: results from the exploration are reproducible, and are clearly exploratory rather than confirmatory. Hypotheses suggested from metboost can then be tested by fitting a single model and testing coefficients in that model in a separate sample. 

The generality and flexibility of metboost comes at the cost of computation time, which also limits the sample size and number of predictors that can be used for the method. Currently, a few thousand predictors and sample sizes <10K are practically feasible in terms of memory use and computation time. This limits applications for high dimensional data from genetics or neuroimaging, which often involve millions of predictors. However, metboost could be feasibly run in a few hours for many data sets in psychology. Computation time also limits further improvements to the method or its interpretation, such as handling multiple grouping variables, obtaining true partial dependence plots, or in obtaining estimates of uncertainty around the predictions. Meta-parameter tuning is critical, but also adds computation time. Our implementation provides easy specification for cross-validated grid tuning for the important metboost meta-parameters, and makes it easy to carry out the tuning in parallel. To carry out the analysis, we suggest utilizing large computing clusters provided by many universities or other cloud computing providers.

An important application for metboost that we left unexplored is for analyses of longitudinal data with many subjects and relatively few measurement occasions. With the data in long form, the unique subject identifier is the grouping variable and time can be included as a predictor in the model, along with other covariates. In this setting, metboost approximates subject specific nonlinear trajectories that are conditional on the included covariates. This approach is similar to using an additive model for time-varying autoregression \citep{bringmann_changing_2016}, but is fully nonparametric and more exploratory. The simulation results carried out in this paper apply to the longitudinal setting by noting that the number of groups is the number of measurement occasions. The results suggest that metboost can have improved prediction error and variable selection performance compared to gbm when the average number of measurement occasions is small, which is often the case in psychology. 

In the future, we hope to extend metboost in several ways. For example, when groups are unknown or appear in the test set but not in training, it may be possible to classify unknown groups by their similarity to groups that are known by clustering. In addition, we hope to generalize metboost for classification or survival outcomes, similar to existing implementations of boosting. Finally, gains in computation time might be realized by using a faster language for scientific computing such as Julia \citep{bezanson_julia:_2017}. 

In conclusion, metboost is a powerful tool for exploratory regression analysis with hierarchically clustered data. We hope that researchers in psychology and other social sciences can use this tool to answer exploratory questions, improve predictions, highlight potential nonlinear effects, and to narrow the number of variables to include in subsequent mixed effects models.

\bibliographystyle{apacite}
\bibliography{lib}     

\end{document}